\documentclass[preprint,review,12pt]{elsarticle}
\usepackage{amssymb}
\usepackage{subcaption}
\usepackage{graphicx}
\usepackage{verbatim}
\usepackage{mathtools}
\usepackage{xfrac}  
\usepackage{hyperref}
\usepackage{mathrsfs}
\usepackage{enumitem}
\usepackage{color}
\usepackage{booktabs}
\usepackage{natbib}
\usepackage[algoruled,boxed,lined]{algorithm2e}
\usepackage{algorithmic}
\usepackage[normalem]{ulem}
\usepackage{multirow}
\usepackage{tabularx}
\usepackage[labelformat=simple]{subcaption}

\def\bf#1{\boldsymbol{#1}}

\def\linesolid{{------}}

\def\colr#1{\textcolor[rgb]{0.635,0.078,0.184}{#1}}  
 
\def\coly#1{\textcolor[rgb]{0.929,0.694,0.125}{#1}} 
\def\colg#1{\textcolor[rgb]{0.466,0.674,0.188}{#1}}  
\def\colb#1{\textcolor[rgb]{0.000,0.4470,0.7410}{#1}}

\def\linesolid {{--------}}

\journal{arXiv}

\begin{document}
	
	\begin{frontmatter}
		
		
		\title{Non-iterative generation of an optimal mesh for a blade passage using deep reinforcement learning}
		
		\author{Innyoung Kim}
%        \ead{innykim@postech.ac.kr}
		\author{Sejin Kim}
%        \ead{ksj908@postech.ac.kr}
		\author{Donghyun You \corref{cor1}}
        \ead{dhyou@postech.ac.kr}
		\address{Department of Mechanical Engineering, Pohang University of Science and Technology, 77 Cheongam-Ro, Nam-Gu, Pohang, Gyeongbuk 37673, South Korea\vspace{-0.4in}} 
		
		\cortext[cor1]{Corresponding author.}
	 
		\begin{abstract}
A method using deep reinforcement learning (DRL) to non-iteratively generate an optimal mesh for an arbitrary blade passage is developed. Despite automation in mesh generation using either an empirical approach or an optimization algorithm, repeated tuning of meshing parameters is still required for a new geometry.
The method developed herein employs a DRL-based multi-condition optimization technique to define optimal meshing parameters as a function of the blade geometry, attaining automation, minimization of human intervention, and computational efficiency.
The meshing parameters are optimized by training an elliptic mesh generator which generates a structured mesh for a blade passage with an arbitrary blade geometry. During each episode of the DRL process, the mesh generator is trained to produce an optimal mesh for a randomly selected blade passage by updating the meshing parameters until the mesh quality, as measured by the ratio of determinants of the Jacobian matrices and the skewness, reaches the highest level. Once the training is completed, the mesh generator create an optimal mesh for a new arbitrary blade passage in a single try without an repetitive process for the parameter tuning for mesh generation from the scratch. The effectiveness and robustness of the proposed method are demonstrated through the generation of meshes for various blade passages.
		\end{abstract}
		
		
	\end{frontmatter}
	
	\section{Introduction}

	A computational mesh is a crucial component for a numerical simulation of fluid flow. The accuracy, stability, and computational cost of the simulation are significantly affected by the quality of the mesh~\cite{zandsalimy2022novel, ali2017optimal}. This is particularly true for simulations around a blade, where aerodynamic losses, pressure drops, and locations of flow separation vary significantly depending on the mesh quality~\cite{moshfeghi2012effects, campos2007numerical}. However, generating high-quality meshes requires significant time and effort. For instance, although a structured mesh is preferred in the simulations of flow around a blade owing to computational accuracy and efficiency~\cite{ali2017optimal, zhang20182d}, achieving high mesh quality is not trivial. For a blade passage in a turbo machine, even if the complex curvature of the blade is present in the computational domain, the cells on the periodic boundaries must be matched with smooth resolution variation. 
	
Research has been conducted to automate the process of generating a mesh using algorithms that define meshing parameters rather than manually specifying the location of individual nodes~\cite{milli2012padram, zagitov2014automatic, costenoble2022automated, marchandise2013cardiovascular, lu2020nnw, zhang20182d, zhu1991new, gargallo2018mesh, zheleznyakova2013molecular}. These algorithms require inputs of meshing parameters such as cell numbers in each coordinate direction, expansion and clustering ratios to generate meshes. While they greatly reduce the time and effort required for mesh generation, the quality of the resulting mesh is heavily dependent on the choice of meshing parameters, often requiring human intervention to adjust parameters for improved mesh quality.

Efforts have been made to reduce human intervention in mesh generation using an optimization technique for determining the meshing parameters.
Dittmer~\cite{dittmer2006mesh}, Ahmad \textit{et al.}~\cite{ahmad2010mesh}, and Islam \textit{et al.}~\cite{islam2015optimization} optimized meshing parameters for meshes around an airfoil, a ground vehicle, and a marine propeller, respectively. By optimizing the meshing parameters, human intervention could be tapered, and optimal meshes could be acquired. However, owing to the nature of optimization, repetition of the mesh generation process are inevitable. Furthermore, even if the optimal parameters are obtained, they are valid only for a specific geometry. The entire optimization process must be repeated from the scratch for a new geometry, which significantly degrades the meshing efficiency. Nevertheless, to the best of our knowledge, there is no method to generate an optimal mesh for a new geometry without an repetitive process for the parameter tuning.
    
    The present study aims at developing a mesh generation method that can generate an optimal mesh for an arbitrary blade passage non-iteratively. To achieve the objective, a multi-condition (MC) optimization method based on deep reinforcement learning (DRL)~\cite{kim2022multi} is employed. Unlike the conventional optimization methods, where optimization is performed under a specific condition, the present DRL-based MC optimization technique produces optimal solutions over a range of conditions. Utilizing the characteristic of MC optimization, parameters used in a meshing algorithm are optimized as a function of the blade geometry such that with the parameters, an optimal mesh is generated in a single try.
    
    The paper is organized as follows: in Section~\ref{sec2}, the present DRL-based MC optimization method is described. In Section~\ref{sec3}, a DRL-based mesh-generation algorithm for a blade passage is developed in the following steps: firstly, an elliptic mesh generator to generate a structured mesh for a blade passage is developed (Section~\ref{sec3.1}), secondly, an MC optimization problem is formulated to determine optimal meshing parameters for the elliptic mesh generator as a function of the blade geometry (Section~\ref{sec3.2}), and finally, training the elliptic mesh generator is performed by solving the MC optimization problem using DRL (Section~\ref{sec3.3}). The capability of the trained mesh generator for optimal mesh generation is examined in Section~\ref{sec4.1}, and its practical applicability is investigated by generating meshes for various blade passages in Section~\ref{sec4.2}. Concluding remarks follow in Section~\ref{sec5}.

	\section{Background} \label{sec2}
	\subsection{Multi-condition optimization} \label{sec2.1}
	The conventional single-condition (SC) optimization problem is defined as follows:
	\begin{equation}
    \begin{gathered}
    \underset{\bf{x}}{\text{max}}f(\bf{x}) \\
    \text{subject to } \bf{x} \in \bf{\Omega},
    \end{gathered}
    \label{eq_sc}
    \end{equation}
    where $\bf{x}$ is a decision vector and $f$ is a real-valued objective function. $\bf{\Omega}$ is the decision space which is defined as the set $\{\bf{x} |~ g_l(\bf{x}) \leq 0, l = 1,2,\ldots,m\}$. $g_l(\bf{x})$ is a constraint function that determines the feasible region of $\bf{x}$, and $m$ is the number of the constraints of $\bf{x}$. The goal of SC optimization is to find an optimal solution $\bf{x}^{*}$ that maximizes $f$.
	
    In contrast to SC optimization, a condition vector $\bf{c}$ and a condition space $\bf{\Phi}$ are introduced in MC optimization. An MC optimization problem is defined as follows:
	\begin{equation}
    \begin{gathered}
    \underset{\bf{x}}{\text{max}}f(\bf{x},\bf{c}) \\
    \text{subject to } \bf{x} \in \bf{\Omega}, \bf{c} \in \bf{\Phi},
    \end{gathered}
    \label{eq_mc}
    \end{equation}
    where $\bf{\Phi}$ is defined as the set $\{\bf{c} |~ h_p(\bf{c}) \leq 0,~p = 1,2,\ldots,q\}$. $h_p(\bf{c})$ is a constraint function that determines the feasible region of $\bf{c}$, and $q$ is the number of the constraints of $\bf{c}$. The goal of an MC optimization is to find optimal solutions $\bf{x}^{*}(\bf{c})$ that maximize $f$ as a function of $\bf{c}$.
	
	\subsection{Deep reinforcement learning} \label{sec2.2}
	DRL is a process of learning an optimal behavior for a complex decision-making process~\cite{sutton2018reinforcement}. 
	In the context of the present study, DRL refers to a process of learning an optimal behavior for determining the meshing parameters as a function of the blade geometry. At each discrete step $t$, an action $\bf{a}_{t} \in \mathcal{A}$ is determined based on a deterministic policy $\bf{\pi} : \mathcal{S} \rightarrow \mathcal{A}$ for a given state $\bf{s}_{t} \in \mathcal{S}$, where $\mathcal{S}$ and $\mathcal{A}$ denote state and action spaces, respectively. After taking the action $\bf{a}_{t}$, a reward $r_{t}(\bf{s}_{t}, \bf{a}_{t}) \in \mathbb{R}$ is obtained, and the next state $\bf{s}_{t+1}$ is determined based on a probability distribution $p(\bf{s}_{t+1} \vert \bf{s}_{t}, \bf{a}_{t}) : \mathcal{S}\times\mathcal{A} \rightarrow \mathcal{P(\mathcal{S})}$, where $\mathcal{P}(\mathcal{S})$ denotes the set of probability distributions over the state space $\mathcal{S}$. The process is repeated until the terminal step $T$, at which point one episode ends. The return $R_{t}$ is defined as the sum of an immediate reward $r_{t}$ and discounted future rewards as follows:
	\begin{equation}
        R_{t} = \sum_{i=t}^{T}\gamma^{(i-t)}~r_{i}(\bf{s}_{i}, \bf{a}_{i}),
        \label{eq_return}
    \end{equation}
	where $\gamma \in [0, 1]$ is a discount factor that determines the weight between short-term and long-term rewards.

	In DRL, the policy $\bf{\pi}_{\bf{\phi}}$ is represented by a nonlinear neural network with network parameters $\bf{\phi}$, i.e. weights and biases, that are adjusted during the training process. The goal of DRL is to find an optimal policy $\bf{\pi}_{\bf{\phi}^{*}}$ that maximizes an objective function $J$ defined as the expectation of the return. Thus, $\bf{\pi}_{\bf{\phi}}$ is updated by $\nabla_{\bf{\phi}}J$ which is calculated using the deterministic policy gradient algorithm~\cite{silver2014deterministic} by applying the chain rule to $J$ as follows:
   \begin{equation}
    	\begin{aligned}
        \nabla_{\bf{\phi}}J & = {\mathop{\mathbb{E}}}_{\bf{s}_{t} \sim p}[\nabla_{\bf{\phi}}Q(\bf{s}_{t}, \bf{a}_{t})] \\
        & = {\mathop{\mathbb{E}}}_{\bf{s}_{t} \sim p}[\nabla_{\bf{a}} Q(\bf{s}_{t},\bf{a}_{t})|_{\bf{a}_{t}=\bf{\pi}_{\bf{\phi}}(\bf{s}_{t})}\nabla_{\bf{\phi}}\bf{\pi}_{\bf{\phi}}(\bf{s}_{t})],
        \label{eq_nabla_objfunction}
        \end{aligned}
    \end{equation}
    where $Q(\bf{s}_{t}, \bf{a}_{t})$ is an action value function defined as follows:
    \begin{equation}
        Q(\bf{s}_{t},\bf{a}_{t}) = {\mathop{\mathbb{E}}}_{\bf{s}_{i > t} \sim p}[R_{t}|\bf{s}_{t}, \bf{a}_{t}].
        \label{eq_Q}
    \end{equation}

    \subsection{Deep-reinforcement-learning-based multi-condition optimization method} \label{sec2.3}
    
    A DRL-based MC optimization method was proposed by Kim \textit{et al.}~\cite{kim2022multi}. The method employs a single-step DRL~\cite{vinquerat2021direct} to address the MC optimization problem. In the single-step DRL, each learning episode consists of a single step. Given a state $\bf{s}_{t}$ at step $t$, an action $\bf{a}_{t}$ is determined, and a reward $r_{t}(\bf{s}_{t}, \bf{a}_{t})$ is obtained. The episode is then terminated without being transitioned to a subsequent state. Since there are no subsequent steps, the return $R_{t}$ and the action value function $Q(\bf{s}_{t}, \bf{a}_{t})$ are defined solely by the immediate reward $r_{t}$ as follows:
    \begin{equation}
        R = Q(\bf{s},\bf{a}) = r(\bf{s}, \bf{a}),
        \label{eq_single_step_return_Q}
    \end{equation}
    where the subscript $t$ is omitted as there is only one step. Using the deterministic policy gradient algorithm in Eq.~\eqref{eq_nabla_objfunction}, learning proceeds to find the optimal policy $\bf{\pi}_{\bf{\phi}^{*}}(\bf{s})$ that maximizes the immediate reward $r$. Thus, by setting the state, action, and reward of the single-step DRL as $\bf{c}$, $\bf{x}$, and $f$, respectively, learning $\bf{x}^{*}$ that maximizes $f$ as a function of $\bf{c}$ becomes possible. Consequently, $\bf{x}^{*}(\bf{c})$, the solution to the MC optimization problem, is obtained.
	
	\section{Deep-reinforcement-learning-based mesh generation method} \label{sec3}
	\subsection{Mesh generation algorithm for a blade passage} \label{sec3.1}
	
	In this section, an elliptic mesh generator to generate a structured mesh for a two-dimensional blade passage is developed. The mesh generator employs an elliptic mesh generation method~\cite{thompson1982elliptic, steger1979automatic} to produce an HOH-type mesh that adopts H-type meshes for the inlet and outlet sides and an O-type mesh near the blade based on the geometric parameters of the blade passage and the meshing parameters. The parameters and descriptions are listed in Table~\ref{table_params}.
	
\begin{figure}
	\centering
	\includegraphics[width=1.0\linewidth]{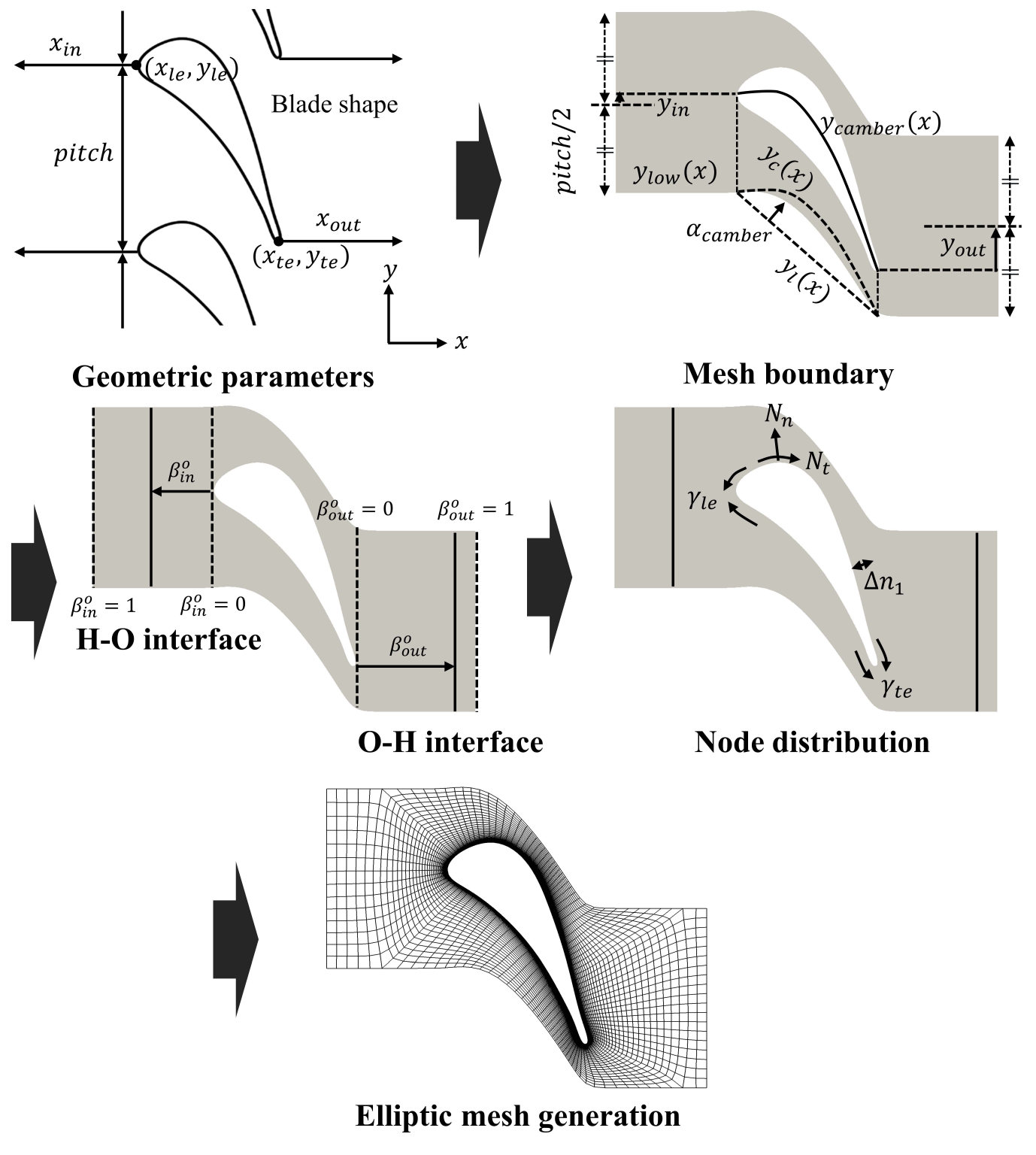}	\caption{Schematic of the elliptic mesh generator that generates a structured mesh of a two-dimensional blade passage.}
	
	\label{fig_mesh_generation_algorithm}
	\end{figure}

    The schematic of the mesh generator is shown in Fig.~\ref{fig_mesh_generation_algorithm}. Firstly, the geometric parameters of the blade passage, including the blade shape, the blade $pitch$, the inlet position $x_{in}$, and the outlet position $x_{out}$, are given to the mesh generator. Subsequently, the mesh boundary is determined based on three meshing parameters $y_{in}$, $y_{out}$, and $\alpha_{camber}$. $y_{in}$ and $y_{out}$ are positions of the inlet and the outlet in the vertical direction, respectively. $\alpha_{camber}$ determines the degree to which the curvature of the lower boundary follows the camber line. The lower boundary $y_{low}(x)$ is defined as follows: 
    \begin{equation} \label{eq_mesh_boundary}
	y_{low}(x) = \left\{ \begin{array}{lll}
	    y_{le} + y_{in} - pitch/2 & x_{le} - x_{in} \leq x < x_{le} , \\
		\alpha_{camber}~y_{c}(x) + (1 - \alpha_{camber})~y_{l}(x)& x_{le} \leq x < x_{te} , \\
		y_{te} + y_{out} - pitch/2 & x_{te} \leq x < x_{te} + x_{out}. \\
		\end{array} \right.
	\end{equation}
    Here, $(x_{le}, y_{le})$ and $(x_{te}, y_{te})$ are the coordinates of the leading and trailing edges, respectively. $y_{c}(x)$ is a scaled camber line in the vertical direction defined as follows:
    \begin{equation}
    y_{c}(x) = \frac{y_{le} + y_{in} - y_{te} - y_{out}}{y_{le} - y_{te}}y_{camber}(x) + \frac{y_{le}y_{out} - y_{te}y_{in}}{y_{le} - y_{te}} - pitch/2,
    \label{eq_scaled_camber}
    \end{equation}
    where $y_{camber}(x)$ is the camber line of the blade. $y_{l}(x)$ is a straight line linearly connecting two points, $(x_{le}, y_{le} + y_{in} - pitch/2)$ and $(x_{te}, y_{te} + y_{out} - pitch/2)$. The upper boundary is determined by shifting the lower boundary vertically by $pitch$ to ensure periodicity. Then, locations of the interface between the O-type mesh and the H-type mesh at the inlet and outlet sides are determined as $x = x_{le} - \beta^{o}_{in}~x_{in}$ and $x = x_{te} + \beta^{o}_{out}~x_{out}$, respectively.
	
\begin{figure}[]
	\centering
	\begin{subfigure}[t]{0.75\linewidth}
	\includegraphics[width=1.0\linewidth]{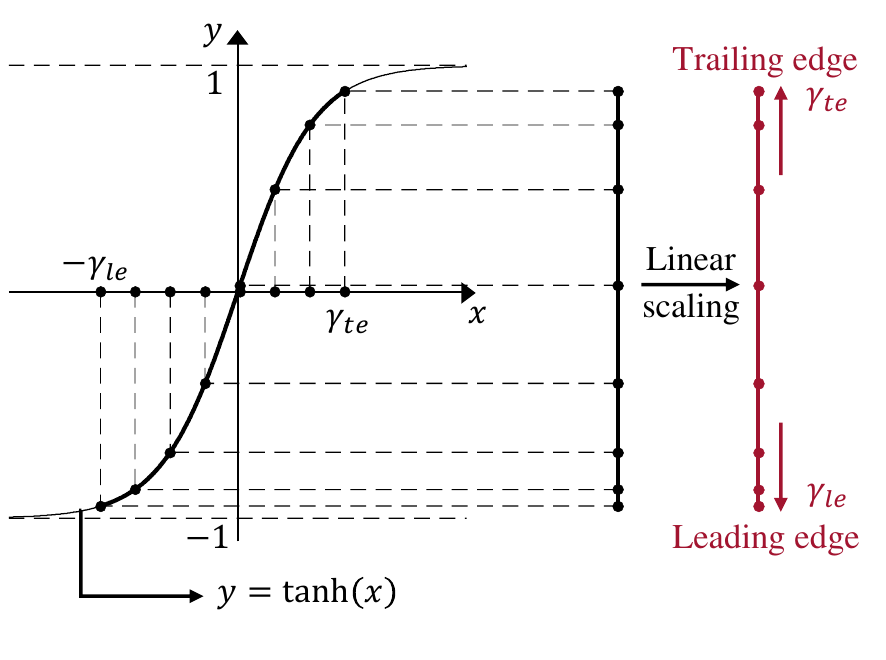}
	\caption{} \label{fig_tanh_a}
	\end{subfigure}
	\begin{subfigure}[t]{0.75\linewidth}
	\includegraphics[width=1.0\linewidth]{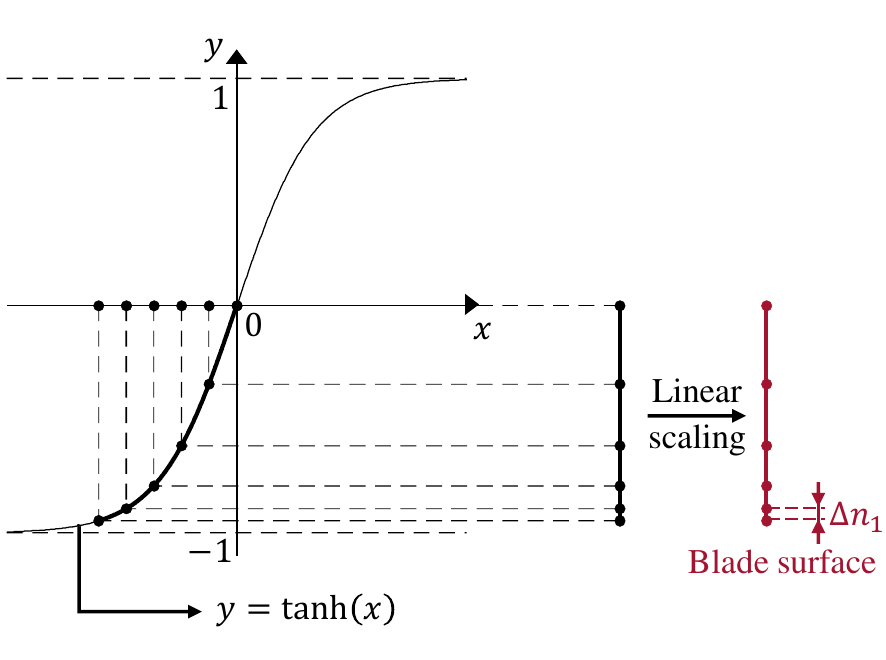}
	\caption{} \label{fig_tanh_b}
	\end{subfigure}
	\caption{Node clustering using a hyperbolic tangent function. Equally distributed points on the $x$-axis are transformed using the hyperbolic tangent function and scaled linearly to the red line. (a) Node clustering at the leading and trailing edges using $\gamma_{le}$ and $\gamma_{te}$. (b) Node clustering to the blade surface using $\Delta n_{1}$.} \label{fig_tanh}
	\end{figure}

	Nodes of the O-type mesh are distributed according to $N_{t}$, $N_{n}$, $\gamma_{le}$, $\gamma_{te}$, and $\Delta n_{1}$. $N_{t}$ and $N_{n}$ are the numbers of nodes of the O-type mesh in tangential and normal directions to the blade surface, respectively. Along the blade surface, $N_{t}$ is divided so that the numbers of nodes on the pressure and suction sides are proportional to their lengths. On each side, the nodes are clustered at the leading and trailing edges for higher resolution, as shown in Fig.~\ref{fig_tanh_a}. Equally distributed nodes are transformed using a hyperbolic tangent function, where $\gamma_{le}$ and $\gamma_{te}$ determine the degrees of clustering at the leading and trailing edges, respectively. The transformed nodes are linearly scaled and distributed along the pressure and suction sides of a blade.
	
	Then, edges are generated by extending the nodes on the blade surface to the outer boundary of the O-type mesh in an outward normal direction. Thereafter, the nodes at the outer boundary are adjusted such that the nodes at the upper and lower boundaries are periodically matched. Along the edge connecting the blade surface to the outer boundary of the O-type mesh, $N_{n}$ nodes are distributed. The nodes are clustered at the blade surface following a hyperbolic tangent function, as shown in Fig.~\ref{fig_tanh_b}. The degree of clustering is determined such that the height of the first cell at the blade surface is equal to the specified height of the first cell $\Delta n_{1}$.
	
	After distributing the nodes of the O-type mesh, an elliptic mesh generation method~\cite{thompson1982elliptic, steger1979automatic} is applied to adjust the node distribution. The elliptic mesh generation method used in the present study is proposed by Hsu and Lee~\cite{hsu1991numerical}. It generates a mesh with high orthogonality near the blade surface without slope discontinuity inside the mesh domain. The method employs Poisson equations defined as follows:
    \begin{equation} \label{eq:poisson_eq_1}
    \frac{\partial^{2} \xi}{\partial x^{2}} + \frac{\partial^{2} \xi}{\partial y^{2}} = P_{1}(\xi,\eta),
    \end{equation}
    \begin{equation} \label{eq:poisson_eq_2}
    \frac{\partial^{2} \eta}{\partial x^{2}} + \frac{\partial^{2} \eta}{\partial y^{2}} = P_{2}(\xi,\eta),
    \end{equation}
    where $(\xi, \eta)$ is the curvilinear coordinate such that the domain of $(x, y)$ is transformed into a rectangular domain. The boundaries of the rectangular domain are denoted by $\xi = 0$, $\xi = \xi_{max}$, $\eta = 0$, and $\eta = \eta_{max}$. The terms $P_{1}$ and $P_{2}$ are control functions that control the node distribution. By interchanging the dependent and independent variables, Eqs.~\eqref{eq:poisson_eq_1} and \eqref{eq:poisson_eq_2} are transformed as follows: 
    \begin{equation} \label{eq:transformed_eq_1}
    A_{1} \frac{\partial^{2} x}{\partial \xi^{2}} - 2A_{2} \frac{\partial^{2} x}{\partial \xi \partial \eta} + A_{3} \frac{\partial^{2} x}{\partial \eta^{2}} = -A_{4}^2(P_{1}\frac{\partial x}{\partial \xi}+P_{2}\frac{\partial x}{\partial \eta}), 
    \end{equation}
    \begin{equation} \label{eq:transformed_eq_2}
    A_{1} \frac{\partial^{2} y}{\partial \xi^{2}} - 2A_{2} \frac{\partial^{2} y}{\partial \xi \partial \eta} + A_{3} \frac{\partial^{2} y}{\partial \eta^{2}} = -A_{4}^2(P_{1}\frac{\partial y}{\partial \xi}+P_{2}\frac{\partial y}{\partial \eta}),
    \end{equation}
    where,
    \begin{equation} \label{eq:transformed_parameters}
    \begin{aligned}
    &A_{1} = (\frac{\partial x}{\partial \eta})^2 + (\frac{\partial y}{\partial \eta})^2,~ A_{2} =\frac{\partial x}{\partial \xi} \frac{\partial x}{\partial \eta} + \frac{\partial y}{\partial \xi} \frac{\partial y}{\partial \eta}, \\
    &A_{3} = (\frac{\partial x}{\partial \xi})^2 + (\frac{\partial y}{\partial \xi})^2,~  A_{4} = \frac{\partial x}{\partial \xi} \frac{\partial y}{\partial \eta} + \frac{\partial x}{\partial \eta} \frac{\partial y}{\partial \xi}.
    \end{aligned}
    \end{equation}
    The values of the control functions at the boundaries $P_{1}(\xi,0)$, $P_{2}(\xi,0)$, $P_{1}(\xi,\eta_{max})$, and $P_{2}(\xi,\eta_{max})$ are determined by applying orthogonality conditions to Eqs.~\eqref{eq:transformed_eq_1} and \eqref{eq:transformed_eq_2}. Then, control functions inside the domain are interpolated based on power-law functions as follows:
    \begin{equation} \label{eq:pq_interpolation_1}
    P_{1}(\xi,\eta) = P_{1}(\xi,0)[1-(\eta/\eta_{max})]^3+P_{1}(\xi,\eta_{max})(\eta/\eta_{max})^3,
    \end{equation}
    \begin{equation} \label{eq:pq_interpolation_2}
    P_{2}(\xi,\eta) = P_{2}(\xi,0)[1-(\eta/\eta_{max})]^3+P_{2}(\xi,\eta_{max})(\eta/\eta_{max})^3.
    \end{equation}
    After generating the O-type mesh using the elliptic generation method, an HOH-type mesh is eventually generated by adding H-type meshes at the inlet and outlet sides. The H-type meshes consist of cells with zero expansion ratio along both horizontal and vertical directions to ensure consistent resolution and minimize the numerical dissipation.

	\subsection{Multi-condition optimization} \label{sec3.2}
	\subsubsection{Mesh quality} \label{sec3.2.1}
	
	In this section, mesh quality that quantitatively evaluates the status of the generated mesh is defined. Generally, mesh quality is evaluated by two factors, the consistency of the spatial distribution and resolution of the cells. Indicators to assess these are \textit{a priori} and \textit{a posteriori} mesh quality, respectively, depending on whether the simulation solution is reflected or not. \textit{A prioiri} mesh quality does not reflect the simulation solution. Instead, it is defined by the geometrical characteristics of the mesh, such as the distortion level and the expansion or compression ratio. Since \textit{a priori} quality can be evaluated before performing the simulation, a lot of efforts have been made to properly define and optimize the quality metrics of a mesh, which is expected to produce more accurate and stable simulation results~\cite{kallinderis2009priori, kallinderis2015priori, garimella2004triangular, fotia2014quality, lowrie2011priori}. The present study employs \textit{a priori} metrics to determine mesh quality.
	
\begin{figure}[]
	\centering
	\begin{subfigure}[t]{1.0\linewidth}
	\includegraphics[width=1.0\linewidth]{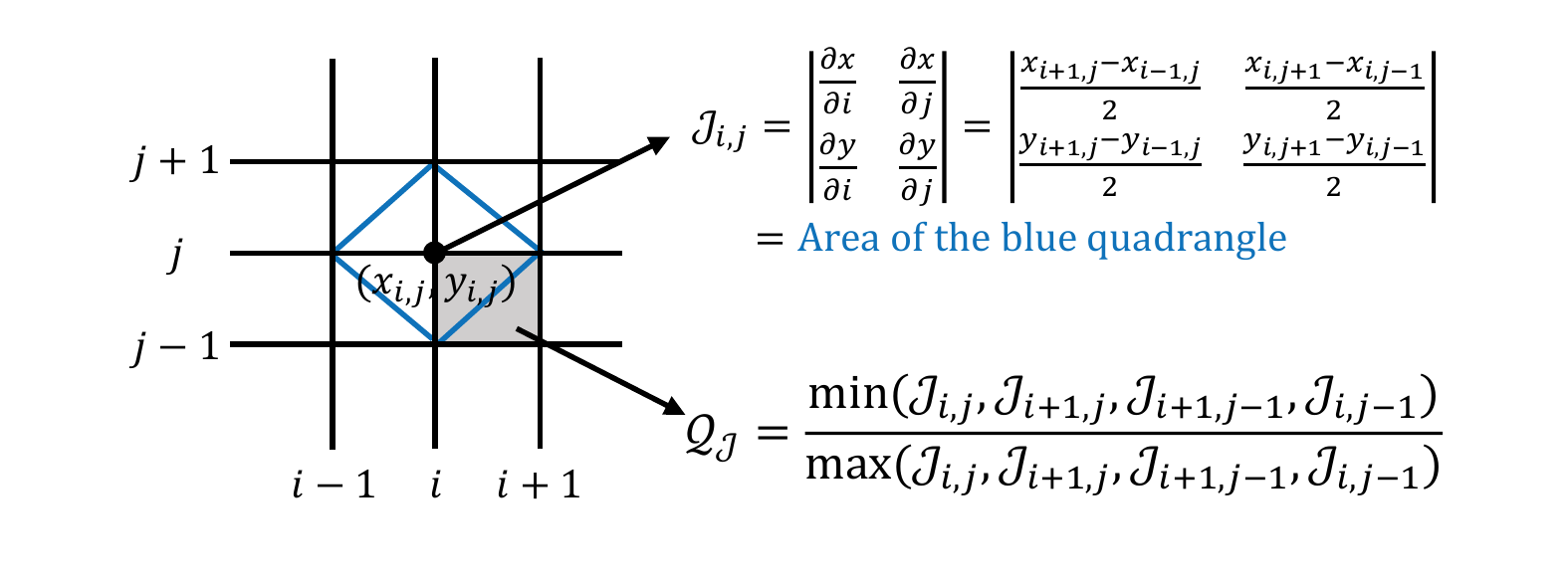}
	\caption{} \label{fig_MeshQuality_a}
	\end{subfigure}
	\begin{subfigure}[t]{1.0\linewidth}
	\includegraphics[width=1.0\linewidth]{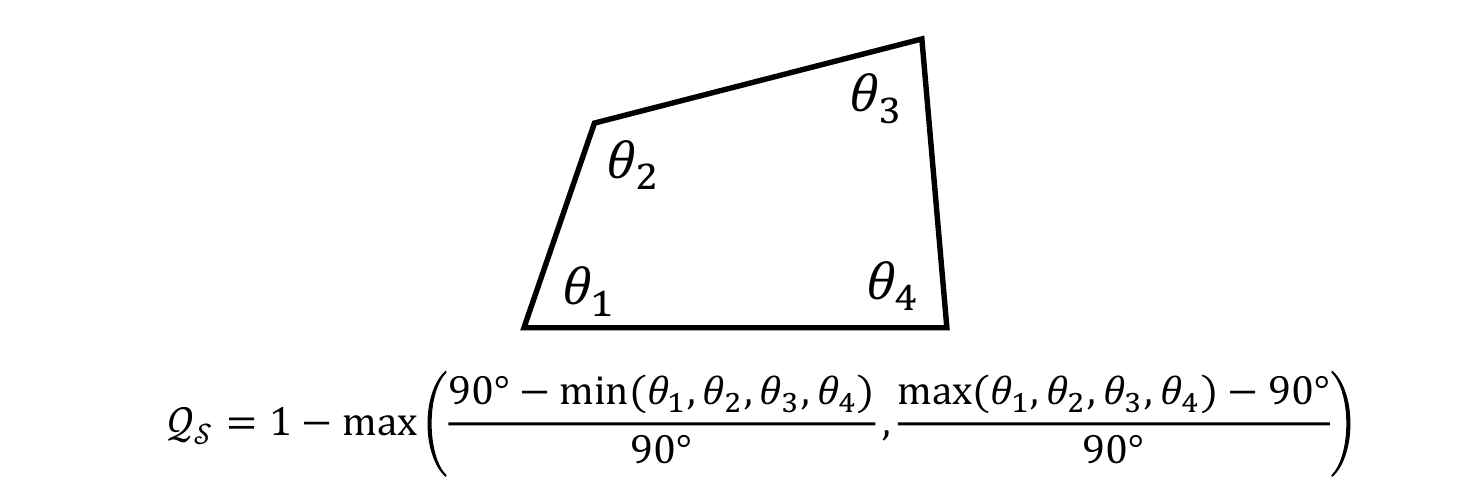}
	\caption{} \label{fig_MeshQuality_b}
	\end{subfigure}
	\caption{Mesh quality metrics. (a) Determinant ratio of the Jacobian matrix $\mathcal{Q}_\mathcal{J}$. (b) Skewness $\mathcal{Q}_\mathcal{S}$.} \label{fig_MeshQuality}
	\end{figure}

	\textit{A priori} mesh quality should be designed to minimize the numerical error caused by geometrical defects of the mesh. To this end, the spatial distribution of the cells should be as uniform as possible while maintaining high orthogonality between adjacent cells. To address the challenges, the ratio of determinants of the Jacobian matrices $\mathcal{Q}_{\mathcal{J}}$ and the skewness $\mathcal{Q}_{\mathcal{S}}$ are considered simultaneously to define mesh quality. Fig.~\ref{fig_MeshQuality} shows how to calculate $\mathcal{Q}_{\mathcal{J}}$ and $\mathcal{Q}_{\mathcal{S}}$ for a cell. $\mathcal{Q}_\mathcal{J}$ is calculated as follows:
	\begin{equation}
	\mathcal{J}_{i,j} = 
	\begin{vmatrix}
	    \frac{\partial x}{\partial i} & \frac{\partial x}{\partial j} \\ 
        \frac{\partial y}{\partial i} & \frac{\partial y}{\partial j} 
	\end{vmatrix} = 
	\begin{vmatrix}
	    \dfrac{x_{i+1,j}-x_{i-1,j}}{2} & \dfrac{x_{i,j+1}-x_{i,j-1}}{2} \\ 
        \dfrac{y_{i+1,j}-y_{i-1,j}}{2} & \dfrac{y_{i,j+1}-y_{i,j-1}}{2} 
	\end{vmatrix}, 
	\label{eq_Jij}
    \end{equation}
    
    \begin{equation}
	\mathcal{Q}_{\mathcal{J}} = \frac{\text{min}(\mathcal{J}_{i,j}, \mathcal{J}_{i+1,j}, \mathcal{J}_{i+1,j-1}, \mathcal{J}_{i,j-1})}{\text{max}(\mathcal{J}_{i,j}, \mathcal{J}_{i+1,j}, \mathcal{J}_{i+1,j-1}, \mathcal{J}_{i,j-1})},
	\label{eq_Qj}
	\end{equation}
	where $x$ and $y$ are the coordinates of the node, and $i$ and $j$ are the indices of the node, as shown in Fig.~\ref{fig_MeshQuality_a}. Firstly, the determinant of the Jacobian matrix $\mathcal{J}_{i,j}$ at each node $(x_{i, j}, y_{i,j})$ is calculated. Subsequently, $\mathcal{Q}_\mathcal{J}$ is calculated by dividing the minimum $\mathcal{J}$ by the maximum $\mathcal{J}$ among the four nodes constituting the cell. As $\mathcal{J}_{i,j}$ is the area of the blue quadrangle, $\mathcal{Q}_\mathcal{J}$ represents the area change among the adjacent cells. A higher $\mathcal{Q}_\mathcal{J}$ value indicates that the area of the cell is more consistent with that of the neighboring cells resulting in a smooth resolution change among the associated cells. 
	
	The skewness $\mathcal{Q}_\mathcal{S}$ is calculated as follows:
	\begin{equation}
	\begin{gathered}
	\mathcal{Q}_\mathcal{S} = 1 - \text{max}(\frac{90^{\circ}-\text{min}(\theta_{1},\theta_{2},\theta_{3},\theta_{4})}{90^{\circ}},\frac{\text{max}(\theta_{1},\theta_{2},\theta_{3},\theta_{4})-90^{\circ}}{90^{\circ}}),
	\end{gathered}
	\end{equation}
	where $\theta$ is the interior angle of the cell as shown in Fig.~\ref{fig_MeshQuality_b}. $\mathcal{Q}_\mathcal{S}$ indicates the distortion of the cell, where a higher $\mathcal{Q}_\mathcal{S}$ value denotes higher orthogonality of the shared edge with the neighboring cells. Since $\mathcal{Q}_\mathcal{J}$ and $\mathcal{Q}_\mathcal{S}$ consider the uniformity of the area distribution of the cells and the orthogonality of the cells, respectively, they are regarded as the representative quality metrics for a quadrilateral mesh~\cite{Gao2017}. Thus, metrics reflecting area changes among adjacent cells or distortion of a cell have been considered crucial factors in the generation of a high-quality quadrilateral mesh, although the exact formulation might be different~\cite{zhang2006adaptive, knupp2000achieving, zhu1991new}.

	The measure of mesh quality $\mathcal{Q}$ is defined as follows:
	\begin{equation}
	\begin{gathered}
	\mathcal{Q} = \frac{(\mathcal{Q}_\mathcal{J})\vert_{min} + (\mathcal{Q}_\mathcal{J})\vert_{avg} + (\mathcal{Q}_\mathcal{S})\vert_{min} + (\mathcal{Q}_\mathcal{S})\vert_{avg}}{4},
	\label{eq_MeshQuality}
	\end{gathered}
	\end{equation}
	where $(~)\vert_{min}$ and $(~)\vert_{avg}$ denote the minimum and average values of all the cells in the O-type mesh, respectively. The minimum values are considered as the simulation can be problematic due to a cell with the lowest quality, although the quality of the other cells is satisfactory. The average values are used to reflect the overall quality distribution. Furthermore, only the cells in the O-type mesh are examined to evaluate the mesh quality. This is because, cells in the O-type mesh are distorted to fit the blade profile, which can degrade mesh quality. On the other hand, the quality of the H-type meshes remains high during the mesh generation process, as most of the cells in the H-type meshes are rectangular with zero expansion ratio. Since the quality of the H-type meshes is higher than that of the O-type mesh, optimization using both types of mesh quality can lead to biased results by increasing the proportions of the H-type meshes. Note that $(\mathcal{Q}_\mathcal{J})\vert_{min}$, $(\mathcal{Q}_\mathcal{J})\vert_{avg}$, $(\mathcal{Q}_\mathcal{S})\vert_{min}$, and $(\mathcal{Q}_\mathcal{S})\vert_{avg}$ exhibit comparable scales, as the values of the metrics lie between 0 and 1 by definition.
 
    For cells at the boundary of the O-type mesh, extra nodes are incorporated across the boundaries to calculate $\mathcal{Q}_\mathcal{J}$. In detail, $\mathcal{J}_{i,j}$ for nodes at the boundaries meeting the H-type meshes are calculated using the coordinates of the adjacent nodes in the H-type meshes. $\mathcal{J}_{i,j}$ for nodes at the periodic boundaries are calculated using the coordinates of the nodes at the opposite periodic boundary as if the same mesh is periodically attached. These treatments enable taking into account smooth resolution changes at the interfaces between the O-type and H-type meshes and at the periodic boundaries.

    \subsubsection{Blade parametrization method} \label{sec3.2.2}
    In the present study, a parametrization method developed by Agromayor \textit{et al.}~\cite{AGROMAYOR2021} is employed to generate various types of blades, such as blades for axial gas turbines, supersonic impulse turbines, and axial compressors. The method generates a two-dimensional blade profile with a continuous curvature using non-uniform rational basis spline curves~\cite{piegl1991nurbs} from blade shape parameters. For the convenience of notation, the blade shape parameters in the method are denoted as a vector $\bf{BSP}$ as follows: 
	\begin{equation}
	\bf{BSP} = (x_{le}, y_{le}, C, \psi, \theta_{le}, \theta_{te}, d_{le}, d_{te}, \rho_{le}, \rho_{te}, t^{u}_{1}, \dots, t^{u}_{k}, t^{l}_{1}, \dots, t^{l}_{k}).
	\end{equation}\label{eq_BSP}

\begin{figure}[]
	\centering
	\begin{subfigure}[t]{0.49\linewidth}
	\includegraphics[width=1.0\linewidth]{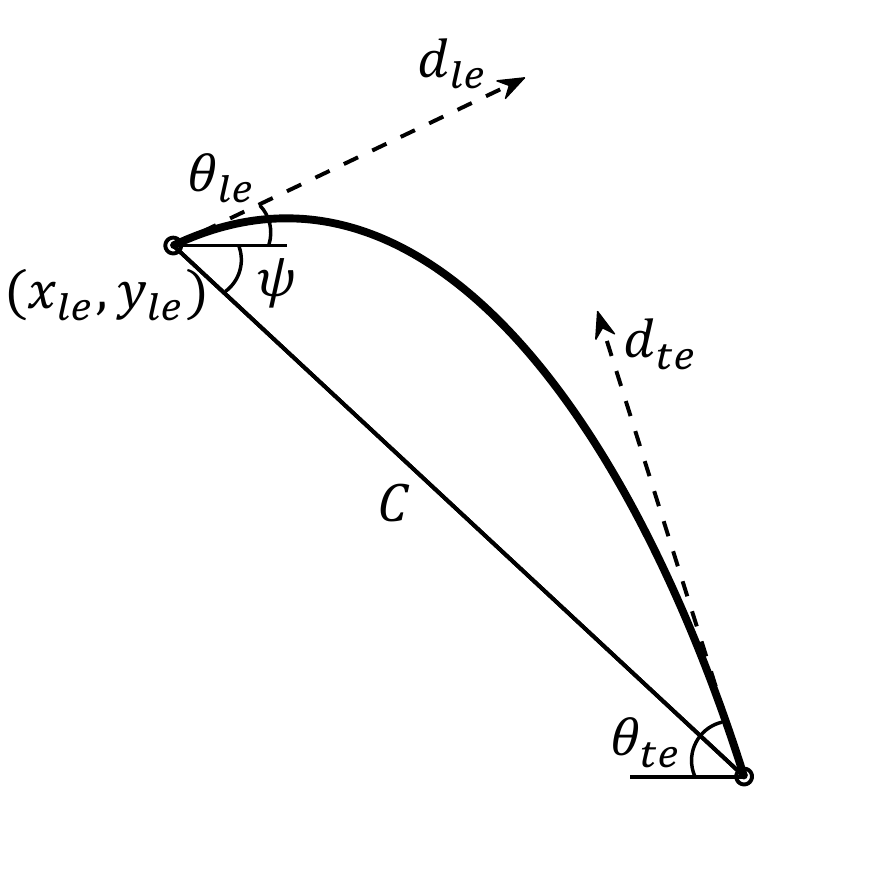}
	\caption{} \label{fig_parablade_a}
	\end{subfigure}
	\begin{subfigure}[t]{0.49\linewidth}
	\includegraphics[width=1.0\linewidth]{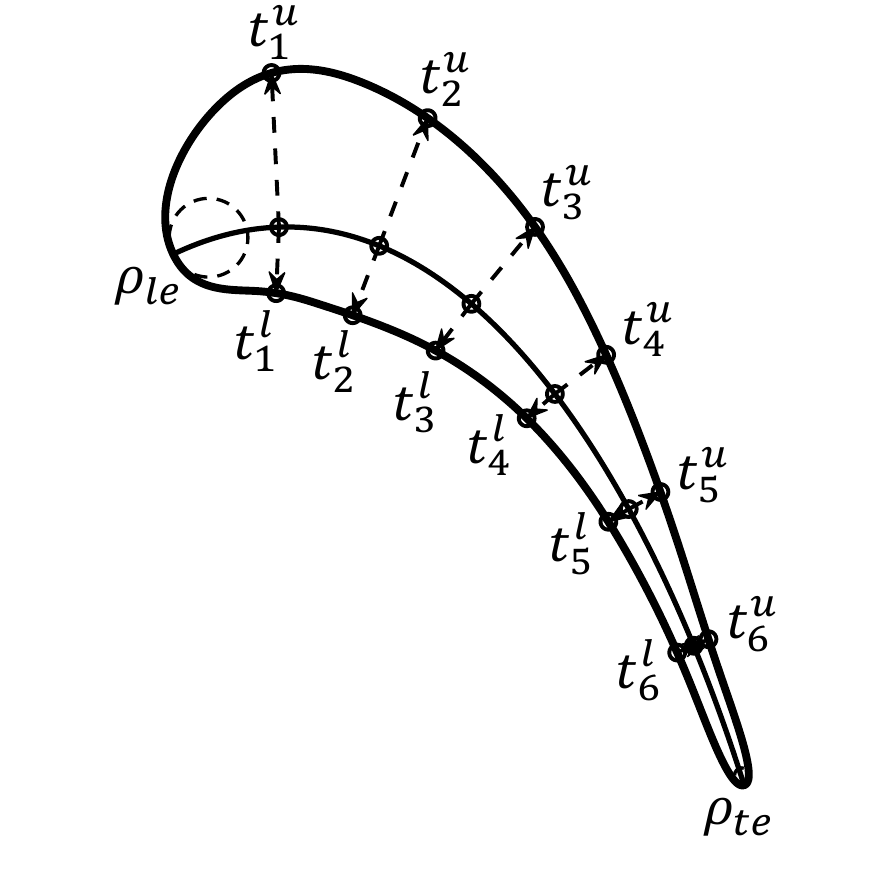}
	\caption{} \label{fig_parablade_b}
	\end{subfigure}
	\caption{Schematics of the blade parametrization method. (a) Camber line construction. (b) Blade profile construction.} \label{fig_parablade}
	\end{figure}

    Fig.~\ref{fig_parablade} illustrates the schematics of the method. Firstly, a camber line is constructed using the first eight variables of $\bf{BSP}$ as depicted in Fig.~\ref{fig_parablade_a}. $C$ is the chord length of the blade, $\psi$ is the stagger angle, $\theta_{le}$ and $\theta_{te}$ are the metal angles of the leading and trailing edges, respectively. $d_{le}$ and $d_{te}$ are the tangent proportions of the leading and trailing edges, respectively. Subsequently, the following variables construct the upper and lower profiles of the blade along the camber line as shown in Fig.~\ref{fig_parablade_b}. $\rho_{le}$ and $\rho_{te}$ are the radii of the curvatures at the leading and trailing edges, respectively. $(t^{u}_{1}, \dots, t^{u}_{k})$ and $(t^{l}_{1}, \dots, t^{l}_{k})$ are the upper and lower thickness distributions, respectively, where $k$ denotes the number of parameters used in the thickness distribution. In the present study, $k = 6$ is used, following the work by Agromayor \textit{et al.}~\cite{AGROMAYOR2021}, as it was confirmed that $k = 6$ is sufficient to represent a diverse range of turbine blade profiles with accuracy comparable to the tolerances used in the manufacturing of blades for axial gas turbines.
    
	\subsubsection{Optimization formulation} \label{sec3.2.3}
    In this section, an MC optimization problem is formulated to obtain the optimal meshing parameters of the elliptic mesh generator according to a wide variety of blade geometries. This capability is particularly crucial when designing high-performance blades, as it requires many systematic simulations with various shapes. In the design process, Reynolds-averaged Navier--Stokes (RANS) simulations are typically employed due to the computational efficiency~\cite{10.1115/1.4001234, HUANG2014481, AGHDASI2022118445}. Therefore, the present study aims to produce meshes suitable for the RANS simulations. The MC optimization problem is defined as follows:\vspace{+0.5in}
	\begin{equation}
    \begin{gathered}
    \underset{\bf{x}}{\text{max}}f(\bf{x},\bf{c}) \\
    \text{subject to } \bf{x} \in \bf{\Omega}, \bf{c} \in \bf{\Phi},
    \end{gathered}
    \label{eq_optform}
    \end{equation}
	where
	{
    
    \centering{$f(\bf{x},\bf{c}) = \mathcal{Q}$,}

    \centering{$\bf{\Omega}=\{(y_{in},y_{out},\alpha_{camber},\beta^{o}_{in},\beta^{o}_{out},N_{t},\gamma_{le},\gamma_{te}) | $}

    \centering{$-0.5C \leq y_{in} \leq 0.5C,~ -0.5C \leq y_{out} \leq 0.5C,~ 0 \leq \alpha_{camber} \leq 1, $}

    \centering{$0.1 \leq \beta^{o}_{in} \leq 0.9,~ 0.1 \leq \beta^{o}_{out} \leq 0.9,~ 100 \leq N_{t} \leq 1000,$}
    
    \centering{$0 \leq \gamma_{le} \leq 5,~ 0 \leq \gamma_{te} \leq 5\}$,}

    \centering{$\bf{\Phi}=\{(\bf{BSP}, pitch, x_{in}, x_{out}, N_{o}, \Delta n_{1}) | $}

    \centering{$\bf{BSP} \in \bf{BSP}_{range},~ 0.3C \leq pitch \leq 1.0C,~ 0 \leq x_{in} \leq C,~ 0 \leq x_{out} \leq C,$}

    \centering{$10000 \leq N_{o} \leq 50000,~ 2\times10^{-5}C \leq \Delta n_{1} \leq 2\times10^{-4}C\}$.}
    
    }
    \noindent $N_{o}=N_{t}\times N_{n}$ is the number of nodes of the O-type mesh. The objective function is mesh quality $\mathcal{Q}$ in Eq.~\eqref{eq_MeshQuality}. The decision vector $\bf{x}$ includes the meshing parameters of the elliptic mesh generator. Ranges of $y_{in}$ and $y_{out}$ are determined such that the inlet and outlet positions can vary vertically within a maximum length of $C$ from the leading and trailing edges, respectively. The proportions of the O-type and H-type meshes at the inlet and the outlet can be adjusted using $\beta_{in}$ and $\beta_{out}$, respectively, ranging from $10\%$ to $90\%$. The minimum value of $N_{t}$ is chosen to provide enough resolution to represent the shape of a blade and the maximum value is determined based on the tangential resolution of the meshes used in the previous RANS simulations around blade passages~\cite{10.1115/1.4001234, HUANG2014481, michelassi2002analysis, zhao2020using}. The range of $\alpha_{camber}$ is set to allow the curvatures of the periodic boundaries to vary from completely following the camber line ($\alpha_{camber} = 1$) to not following the camber line ($\alpha_{camber} = 0$). The distribution of the nodes along the blade surface can be adjusted from uniform spacing to a higher degree of clustering toward the leading and trailing edges by varying $\gamma_{le}$ and $\gamma_{te}$, respectively.

    The condition vector $\bf{c}$ is designated to consider both the blade geometry and the flow condition. The ranges of the blade shape parameters, denoted as $\bf{BSP}_{range}$, are established to encompass the complete feasible domain of each parameter~\cite{AGROMAYOR2021} to generate a diverse set of blade shapes, excluding profiles that are not simply connected or have more than two extreme points. Along with the blade shape, the geometric parameters such as $pitch$, $x_{in}$, and $x_{out}$ are included in $\bf{c}$, and their ranges are defined to cover the values utilized in the prior studies of blade passages~\cite{arts1990aero, fransson1993panel, stadtmuller2001test, anand2020adjoint}. $N_{o}$, which determines resolution of the entire mesh, and $\Delta n_{1}$ which determines resolution near the blade surface are incorporated in $\bf{c}$ to consider the flow condition. The range of $N_{o}$ is set based on the number of cells used in the prior RANS simulations~\cite{10.1115/1.4001234, marciniak2010predicting, michelassi2002analysis, zhao2020using}. The range of $\Delta n_{1}$ is set to satisfy a unity wall resolution calculated by the flat-plate boundary layer theory~\cite{schlichting1961boundary} at $10^{5} \leq Re \leq 10^{6}$, where $Re$ is the Reynolds number. By solving the MC optimization problem, the optimal meshing parameters that maximize mesh quality are obtained as a function of $\bf{c}$. 
 
	\subsection{Deep-reinforcement-learning-based mesh generation algorithm} \label{sec3.3}
	The MC optimization problem is solved using DRL to train the mesh generator such that it can generate an optimal mesh for various blade passages without iteration. For DRL, the actor-critic method~\cite{konda2000actor} is employed. An actor $\bf{\pi}_{\bf{\phi}}$ and a critic $Q_{\bf{\zeta}}$ are functions parameterized using nonlinear neural networks. The network parameters $\bf{\phi}$ and $\bf{\zeta}$ correspond to the weights and the biases of the actor and critic networks, respectively. The actor network $\bf{\pi}_{\bf{\phi}} (\bf{s})$ determines an optimal action that maximizes the expectation of the return in Eq.~\eqref{eq_return} according to the given state. The critic network $Q_{\bf{\zeta}} (\bf{s}, \bf{a})$ predicts the action value function in Eq.~\eqref{eq_Q} depending on the state and the action. Note that since the single-step DRL is employed, the return is identical to the reward. The state $\bf{s}$, the action $\bf{a}$, and the reward $r$ of DRL are defined as $\bf{c}$, $\bf{x}$, and $\mathcal{Q}$, respectively. To ensure consistency in scaling across different components, all variables in $\bf{s}$ and $\bf{a}$ are normalized to the range of $-1$ to $1$ based on the minimum and maximum values of each component defined in Eq.~\eqref{eq_optform}.

    The training procedure for the mesh generator using DRL is presented in Algorithm~\ref{alg_drl}. For every episode, a state $\bf{s}$ is given by random sampling of $\bf{c} = (\bf{BSP}, pitch, x_{in}, x_{out}, N_{o}, \Delta n_{1})$ from $\bf{\Phi}$, and the actor determines an action as $\bf{a} = \bf{\pi}_{\bf{\phi}} (\bf{s})$ with exploration noise $\bf{\epsilon}$. After that, the elliptic mesh generator produces a mesh using meshing parameters specified as the action. Thereafter, mesh quality $\mathcal{Q}$ of the generated mesh is evaluated, and the reward $r$ is obtained. The data of $(\bf{s}, \bf{a}, r)$ is then stored in a buffer. Based on the data in the buffer, the critic network is updated to predict the reward more accurately, and the actor network is updated to produce the action that maximizes the reward, and the episode terminates. This process is repeated until the convergence of the networks. Note that as one episode is composed of a single step $(\bf{s}, \bf{a}, r)$, the converged mesh generator is able to generate a mesh yielding the maximal reward at a single attempt.

    The network structures and the hyperparameters are set according to Kim \textit{et al.}~\cite{kim2022multi}. This setting has advantages of avoiding local minima and finding the optimal solutions precisely for optimization problems with nonlinear characteristics. Following the work of Kim \textit{et al.}~\cite{kim2022multi}, the actor and the critic are parameterized as fully-connected networks with four hidden layers of $512$, $256$, $256$, and $128$ neurons. The hidden layers in both networks use the Leaky ReLU activation function~\cite{maas2013rectifier}. The output layer of the actor network employs the hyperbolic tangent activation function, allowing the action values to range from $-1$ to $1$. The network parameters are updated using the Adam optimizer~\cite{kingma2017adam} with a learning rate of $10^{-4}$ and a mini-batch size $N_b$ of $100$, which is a commonly used optimizer in actor-critic algorithms~\cite{lillicrap2019, RN363}. The exploration noise $\bf{\epsilon}$ is determined from a normal distribution $\mathcal{N}(0,\sigma^{2})$, with a mean of $0$ and the standard deviation $\sigma$, which is set as follows:
    
	\begin{equation} \label{eq_expnoise}
	\sigma = \left\{ \begin{array}{ll}
	    1 & \textrm{episode } \leq 1000, \\
		0.25(\text{cos}(\frac{2\pi}{1000}\times\text{episode})+1) & \textrm{episode } > 1000. \\
		\end{array} \right.
	\end{equation}
    Large $\sigma$ in early episodes allows gathering of various data. Then, a cosine function is employed to balance exploration and exploitation by periodically changing the magnitude of the noise. The updating frequency of the actor network is set to every two episodes, and that of the critic network is set to every episode for stability in the learning process~\cite{RN363}.

\begin{figure}[]
	\centering
	\centerline{\includegraphics[width=0.8\linewidth]{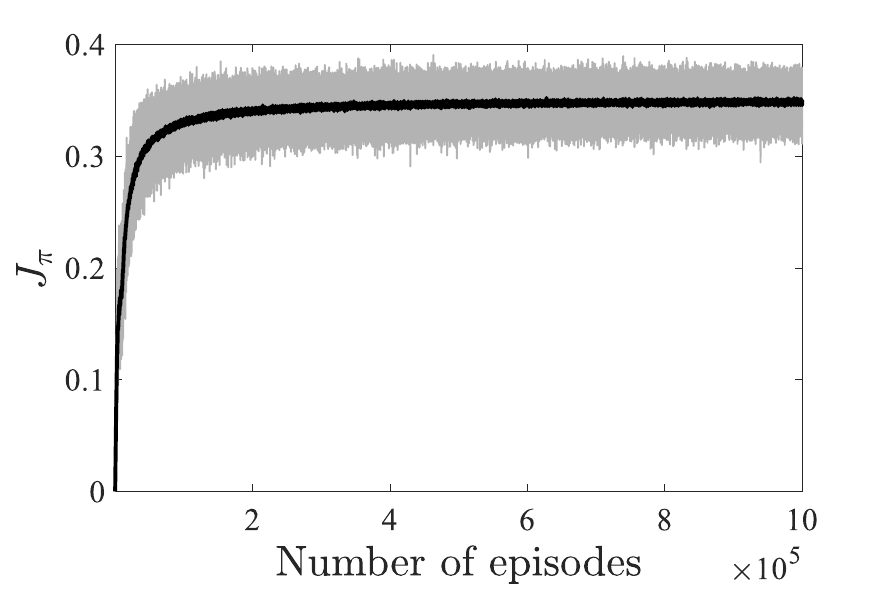}}
	\captionsetup{width=1.0\linewidth}
	\caption{The loss of the actor network $J_{\bf{\pi}}$ as a function of the number of episodes. The grey line indicates the instantaneous loss, and the black line represents the moving average of the loss over 100 episodes.} \label{fig_actor_conv}
    \end{figure}

    Fig.~\ref{fig_actor_conv} shows $J_{\bf{\pi}}$, the loss of the actor network, as a function of the number of episodes. For every episode, $N_b$ data of $(\bf{s}, \bf{a})$ are randomly sampled from the buffer, and $J_{\bf{\pi}}$ is calculated as follows:
    
    \begin{equation}
    J_{\bf{\pi}} = \frac{1}{N_b}\sum_{i=1}^{N_{b}} Q_{\bf{\zeta}}(\bf{s}_{i}, \bf{a}_{i}).
    \label{eq_actor_loss}
    \end{equation}
    In early episodes, low values of $J_{\bf{\pi}}$ are observed as the network is insufficiently trained to determine meshing parameters adequately. As the number of episodes increases, the network is gradually updated to produce higher-quality meshes for a newly given blade passage at each episode. Consequently, high-reward data accumulates in the buffer, and the value of $J_{\bf{\pi}}$ increases and eventually converges. For the present optimization problem, it is found that about $10^{6}$ episodes are necessary to make the network be sufficiently trained.

    \section{Results and discussion} \label{sec4}
    \subsection{Generation of a mesh with optimal quality} \label{sec4.1}
    In this section, the capability of the trained mesh generator to generate an optimal-quality mesh is examined. To this end, the quality of meshes generated without iteration by the present method is compared with that obtained by the conventional optimization method, which requires iterations for mesh generation. From the comparative analysis, whether the quality generated by the present method is optimal is determined. Additionally, the number of iterations required to attain the optimal quality by the conventional approach is examined to identify the computational efficiency of the present method. 
    
    For comparison, four test conditions in the condition space are defined using the existing turbomachinery blades. Four blades, LS89~\cite{arts1990aero}, STD10~\cite{fransson1993panel}, T106A~\cite{stadtmuller2001test}, and SIRT~\cite{anand2020adjoint}, which have different geometries and applications, are selected to represent various blade geometries. LS89 and T106A are axial gas turbine blades for high and low pressure, respectively. STD10 is an axial compressor blade, and SIRT is a blade of a supersonic impulse turbine. For the four blades, the test conditions are designated as follows:
    \begin{equation}
    \begin{gathered}
    \bf{c}_{\text{LS89}} = (\bf{BSP}_{\text{LS89}},~ pitch_{\text{LS89}},~ x_{in} = 0.5C,~ x_{out} = 0.5C,\\ 
    N_{o} = 30000,~ \Delta n_{1} = 1.1\times10^{-4}C), \\
    \bf{c}_{\text{STD10}} = (\bf{BSP}_{\text{STD10}},~ pitch_{\text{STD10}},~ x_{in} = 0.5C,~ x_{out} = 0.5C,\\ 
    N_{o} = 30000,~ \Delta n_{1} = 1.1\times10^{-4}C), \\
    \bf{c}_{\text{T106A}} = (\bf{BSP}_{\text{T106A}},~ pitch_{\text{T106A}},~ x_{in} = 0.5C,~ x_{out} = 0.5C,\\ 
    N_{o} = 30000,~ \Delta n_{1} = 1.1\times10^{-4}C), \\
    \bf{c}_{\text{SIRT}} = (\bf{BSP}_{\text{SIRT}},~ pitch_{\text{SIRT}},~ x_{in} = 0.5C,~ x_{out} = 0.5C,\\ 
    N_{o} = 30000,~ \Delta n_{1} = 1.1\times10^{-4}C). 
    \end{gathered}
    \label{eq_test_condition}
    \end{equation}
    The values of $x_{in}$, $x_{out}$, $N_{o}$, and $\Delta n_{1}$ in the test conditions are the medians of each variable range in the condition space. Note that no special treatment is applied to the test conditions to guarantee that they are treated equally to other conditions in the condition space, thereby making the test conditions unbiased for the analysis.

    The comparative analysis is conducted in the following manner. Firstly, the quality of a mesh generated by the present DRL-based mesh generator at a single attempt, denoted as $\mathcal{Q}_{SA}$, is evaluated for each test condition. Subsequently, iterative optimization is performed to maximize the normalized mesh quality $\mathcal{Q}/\mathcal{Q}_{SA}$ for each test condition. Iterative optimization is conducted by fixing the condition in Algorithm~\ref{alg_drl} and performed $10$ times with different random seeds to minimize stochastic impact due to exploration of DRL and enhance the possibility of finding the global optimum. 
    
\begin{figure}[]
	\centering
	\centerline{\includegraphics[width=0.8\linewidth]{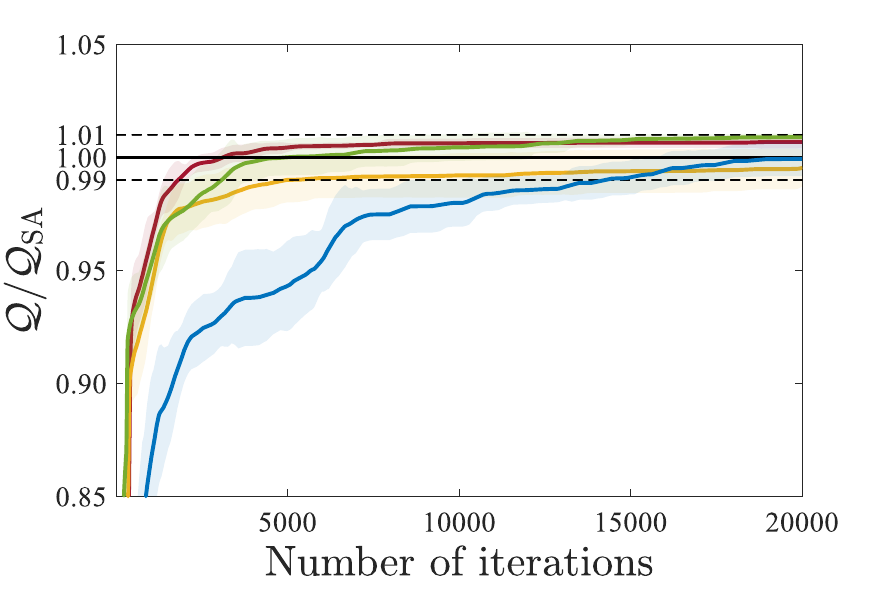}}
	\captionsetup{width=1.0\linewidth}
	\caption{Mesh quality $\mathcal{Q}$ as a function of the number of iterations for the test conditions ($\bf{c}_{\text{LS89}}$, $\bf{c}_{\text{STD10}}$, $\bf{c}_{\text{T106A}}$, and $\bf{c}_{\text{SIRT}}$) by iterative optimization from the scratch. \colr{\linesolid}, $\bf{c}_{\text{LS89}}$; \coly{\linesolid}, $\bf{c}_{\text{STD10}}$; \colg{\linesolid}, $\bf{c}_{\text{T106A}}$; \colb{\linesolid}, $\bf{c}_{\text{SIRT}}$. Each bold line indicates the average of 10 independent runs with different random seeds, and the shaded area represents the standard deviation. Note that $\mathcal{Q}$ is normalized by $\mathcal{Q}_{\text{SA}}$, the quality of a mesh obtained at a single attempt by the trained mesh generator using deep reinforcement learning (DRL), for each test condition.} \label{fig_iter_comp}
	\end{figure}

\begin{figure}[]
	\centering
	\centerline{\includegraphics[width=0.95\linewidth]{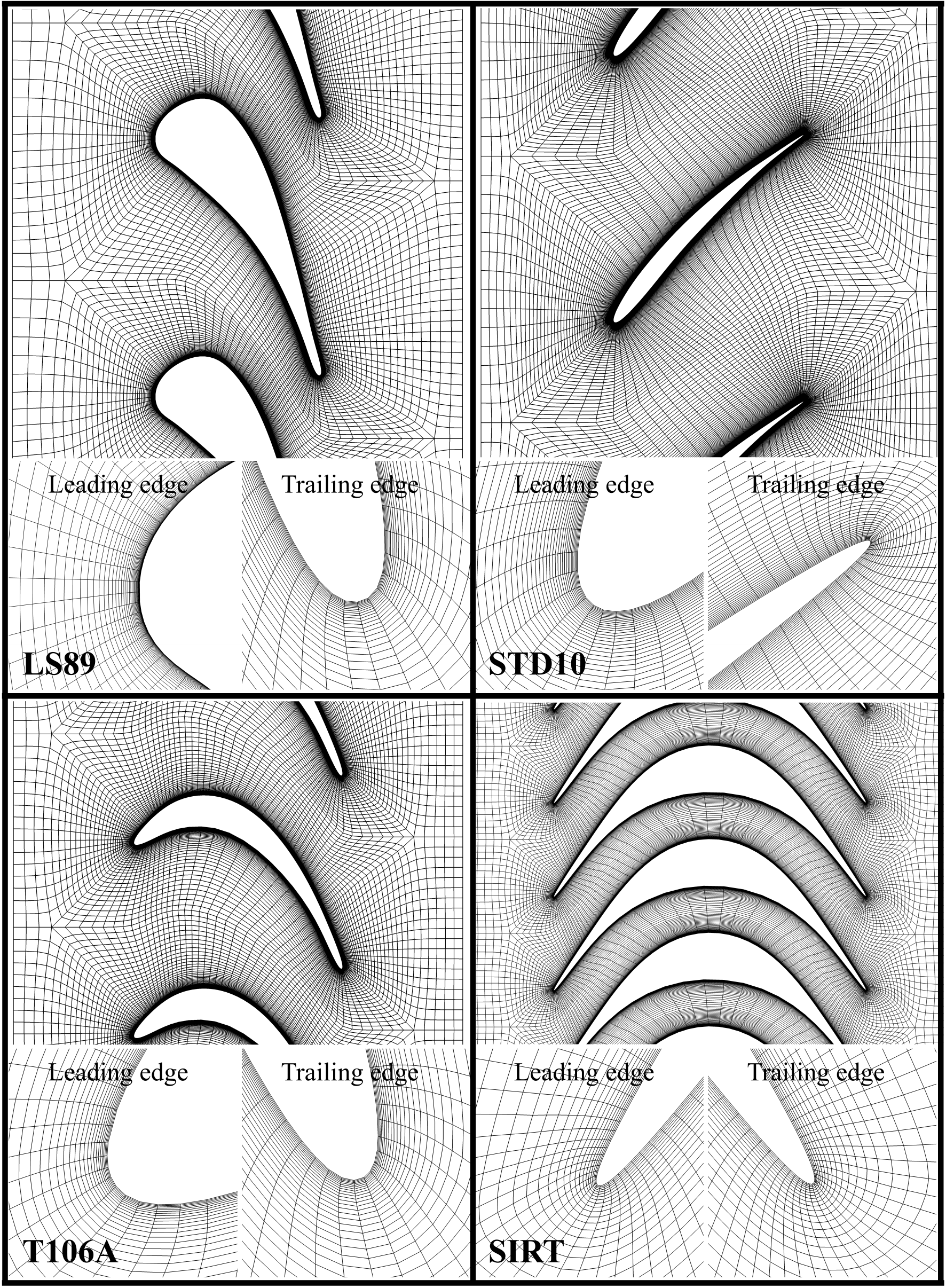}}
	\captionsetup{width=1.0\linewidth}
	\caption{Optimal meshes non-iteratively generated by the trained DRL-based mesh generator for the test conditions. Each section is composed of a full view of the mesh, magnified views around the leading and trailing edges. Every 2nd line is shown for clarity.} \label{fig_testset}
	\end{figure}

    Fig.~\ref{fig_iter_comp} shows the results of iterative optimization for the four test conditions. As the number of iterations progresses, the averaged $\mathcal{Q}/\mathcal{Q}_{SA}$ increases and shows convergence around the $10^{4}$th iteration for $\bf{c}_{\text{LS89}}$, $\bf{c}_{\text{STD10}}$, and $\bf{c}_{\text{T106A}}$, while $\bf{c}_{\text{SIRT}}$ shows convergence around the $2\times10^{4}$th iteration. This is because $\bf{c}_{\text{SIRT}}$ consists of a highly cambered blade shape with a small $pitch$ as shown in Fig.~\ref{fig_testset}, and therefore, more challenging for mesh generation. The magnitudes of the standard deviation gradually decrease as the averaged values converge, indicating that the optimization results are consistent across different random seeds, and the converged values are expected to guarantee optimality. The converged values lie between $0.99$ and $1.01$ for all test conditions, confirming that the quality of the meshes acquired by the present DRL-based mesh generator at a single attempt are comparable to the optimal quality obtained iteratively from the scratch. Furthermore, considering about $5000$ to $10000$ iterations are required to achieve the normalized mesh quality of $1$ under the test conditions, it is expected that about an order of $10^{3}\textup{--}10^{4}$ iterations are necessary to optimize a mesh from the scratch for a new blade passage. Therefore, the present DRL-based mesh generator is computationally extremely efficient as only a single trial is necessary.

    Fig.~\ref{fig_testset} illustrates the optimal meshes produced at a single attempt by the present mesh generator for the four test conditions. The values of the quality metrics are listed in Table~\ref{table_testset_quality}. Higher values represent higher quality, where the maximum value is $1$. $(\mathcal{Q}_\mathcal{J})\vert_{avg}$ which indicates the overall consistency of resolution changes across cells, exhibits values of $0.90$ to $0.92$, as depicted in Fig.~\ref{fig_testset}, where smooth changes in resolution are observed within the domains. Note that the cells in the periodic boundaries are perfectly matched and exhibit smooth transition of mesh resolution across the boundaries. The boundaries between the O-type and H-type meshes also exhibit smooth changes in resolution. These smooth resolution changes at the boundaries are owing to the additional treatments applied in the calculation of $\mathcal{Q}_{\mathcal{J}}$ at the boundaries of the O-type mesh. The calculation involves incorporation of extra nodes across the boundaries to ensure that the consistency of the resolution changes at the boundaries is considered in the optimization process. $(\mathcal{Q}_\mathcal{S})\vert_{avg}$ represents the orthogonality of cells and exhibits values of $0.80$ to $0.86$, where $0.80$ corresponds to an angle of distortion of $18^{\circ}$. This is identified in Fig.~\ref{fig_testset}, where cells with high orthogonality near the blade surface, including the leading and trailing edges, and within the domain are observed. Note that the minimum values of $\mathcal{Q}_{\mathcal{J}}$ and $\mathcal{Q}_{\mathcal{S}}$ are included in optimization to reduce the risk of the simulation being unstable by the cells with the lowest quality, even if the quality of the remaining cells is satisfactory. The minimum values of $\mathcal{Q}_{\mathcal{J}}$ and $\mathcal{Q}_{\mathcal{S}}$ are $63.74\%$ to $68.89\%$ and $54.65\%$ to $61.25\%$ of their average values, respectively, except for $\mathcal{Q}_{\mathcal{S}}$ for $\textbf{c}_{\text{LS89}}$. $(\mathcal{Q}_\mathcal{S})\vert_{min}$ for $\textbf{c}_{\text{LS89}}$ is $40.22\%$ of its corresponding average value due to the highest stagger angle of $55.0^{\circ}$ among the test conditions.
    
	\subsection{Mesh generation for untrained blade passages} \label{sec4.2}
\begin{figure}[]
	\centering
	\centerline{\includegraphics[width=1.0\linewidth]{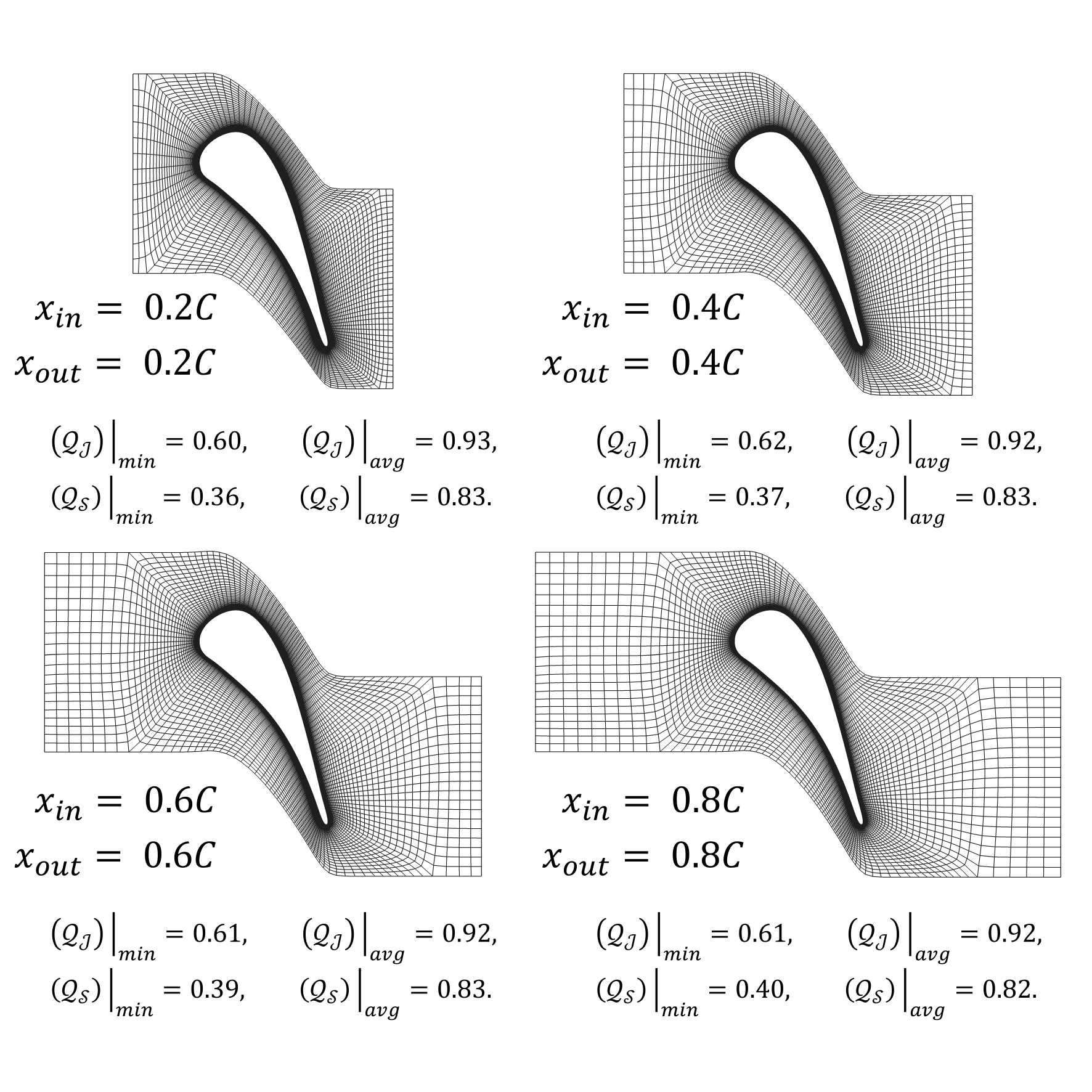}}
	\captionsetup{width=1.0\linewidth}
	\caption{Optimal meshes non-iteratively generated by the trained DRL-based mesh generator and the values of the quality metrics as a function of $x_{in}$ and $x_{out}$. Every 2nd line is shown for clarity.} \label{fig_vary_inout}
	\end{figure}

\begin{figure}[]
	\centering
	\captionsetup{width=1.0\linewidth}
	\centerline{\includegraphics[width=1.0\linewidth]{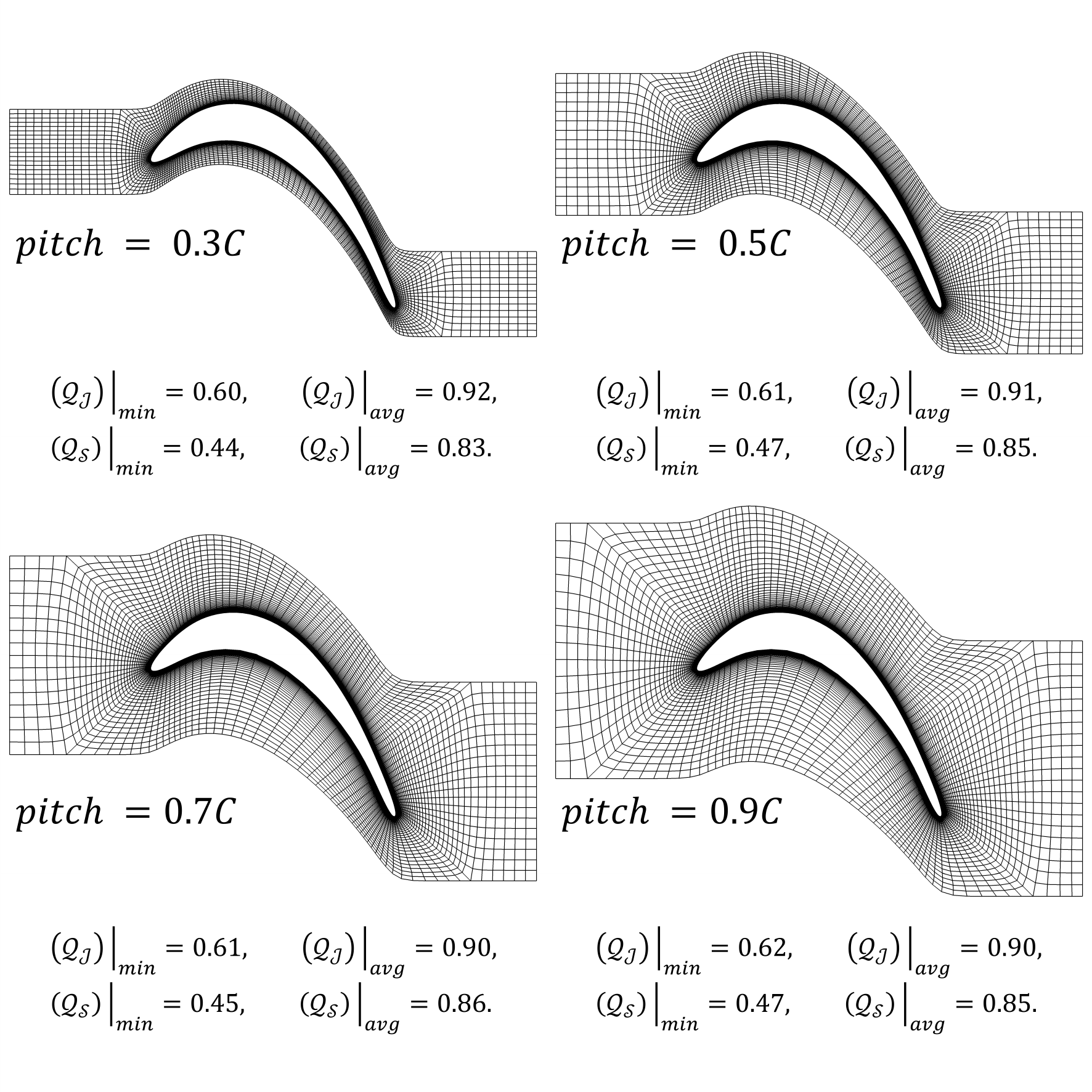}}
	\caption{Optimal meshes non-iteratively generated by the trained DRL-based mesh generator and the values of the quality metrics as a function of $pitch$. Every 2nd line is shown for clarity.} \label{fig_vary_pitch}
	\end{figure}

\begin{figure}[]
	\centering
	\centerline{\includegraphics[width=1.0\linewidth]{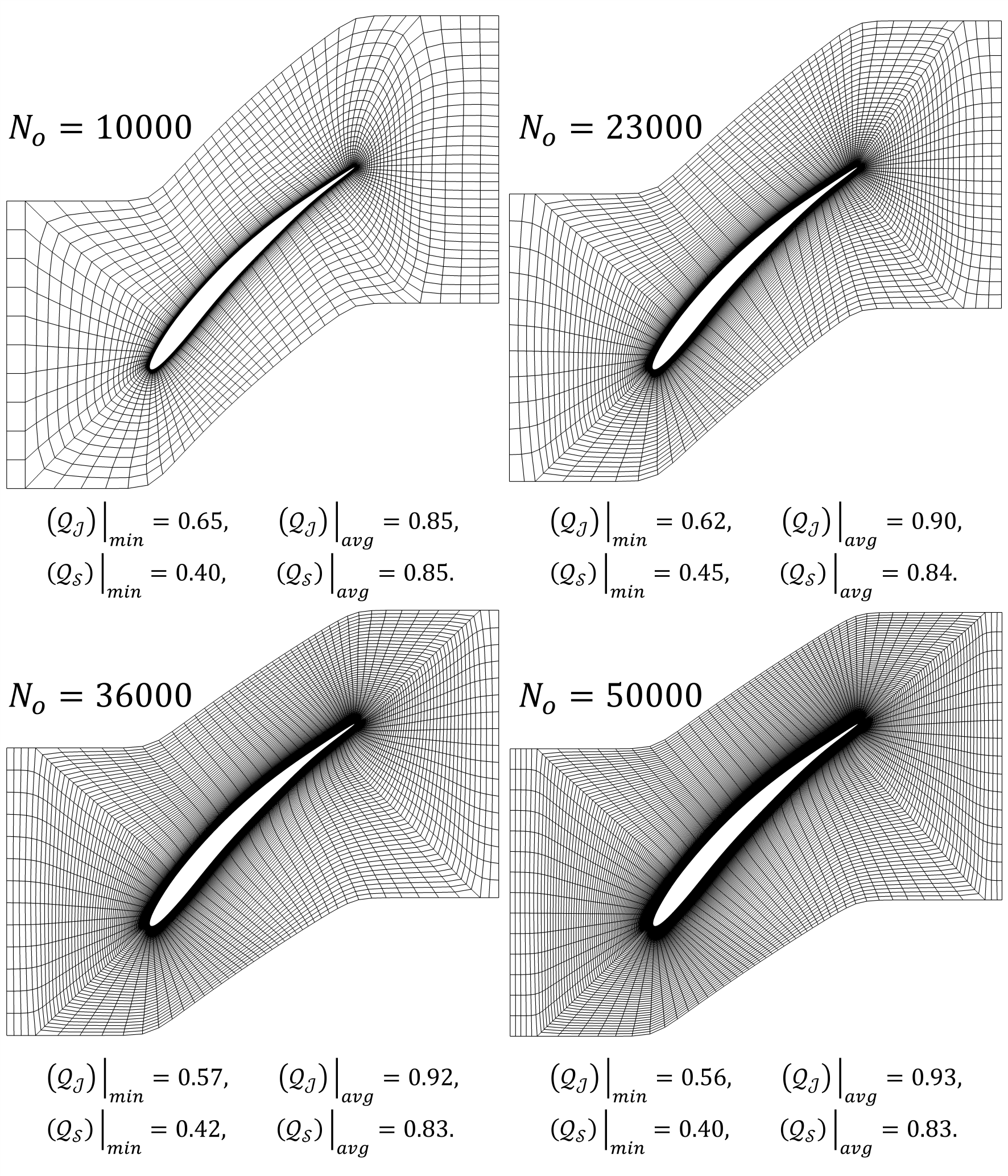}}
	\captionsetup{width=1.0\linewidth}
	\caption{Optimal meshes non-iteratively generated by the trained DRL-based mesh generator and the values of the quality metrics as a function of $N_{o}$. Every 2nd line is shown for clarity.} \label{fig_vary_No}
	\end{figure}

\begin{figure}[]
	\centering
	\centerline{\includegraphics[width=1.0\linewidth]{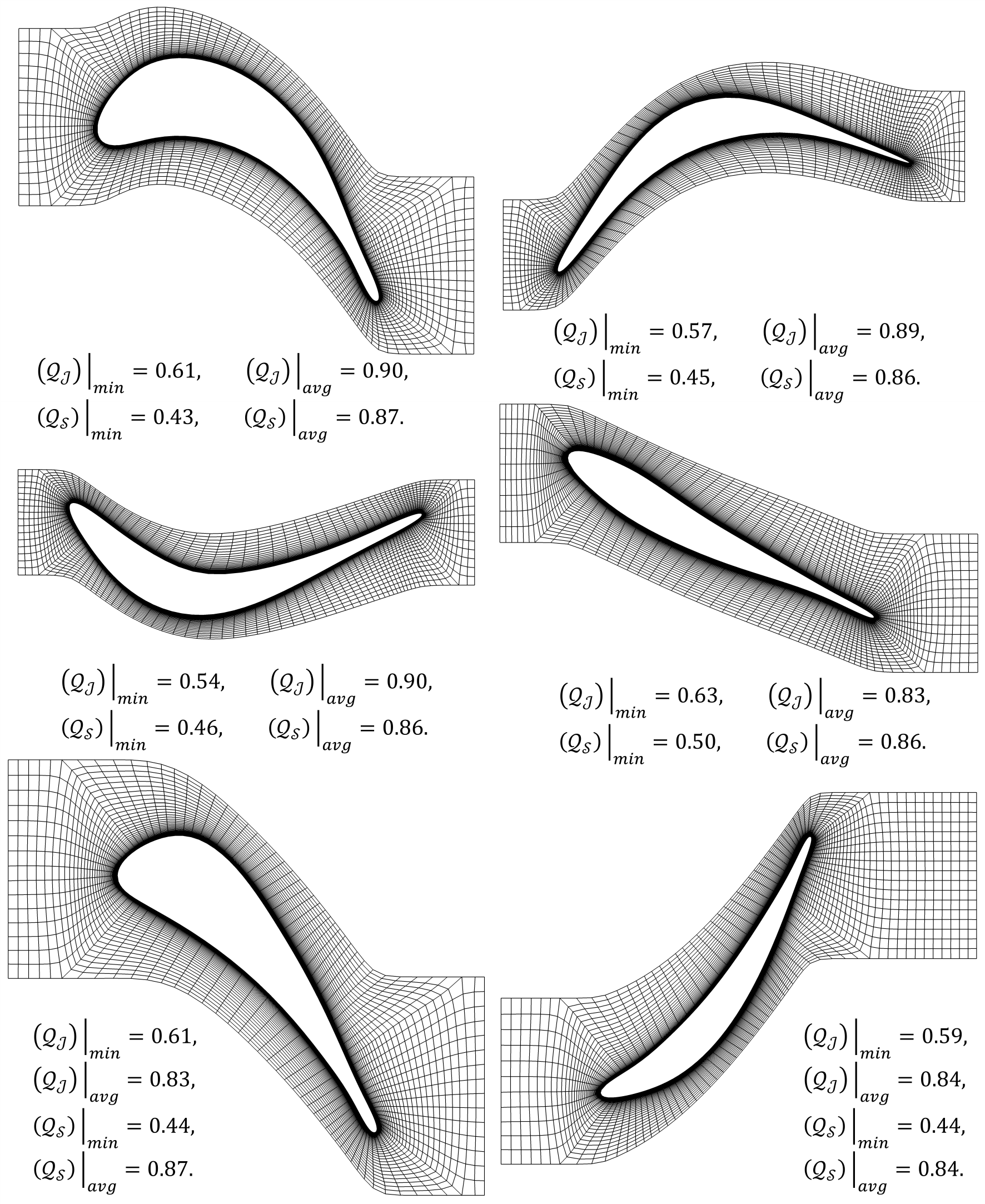}}
	\captionsetup{width=1.0\linewidth}
	\caption{Optimal meshes non-iteratively generated by the trained DRL-based mesh generator and the values of the quality metrics for various blade passages. Every 2nd line is shown for clarity.} \label{fig_arb}
	\end{figure}

    In situations where the geometry undergoes frequent modifications, such as during the design process, efficiently generating meshes for various blade passages is necessary. To identify the practical applicability of the trained mesh generator, meshes are generated for variations of a blade passage and other passages with arbitrary blade shapes. For the convenience of notation, the condition vector $\bf{c} = (\bf{BSP}, pitch, x_{in}, x_{out}, N_{o}, \Delta n_{1})$ in Eq.~\eqref{eq_optform} is used to express blade passages for mesh generation by changing each component of $\bf{c}$. Figs.~\ref{fig_vary_inout},~\ref{fig_vary_pitch},~\ref{fig_vary_No}, and~\ref{fig_arb}, illustrate meshes generated by the trained mesh generator at a single attempt, along with the values of the quality metrics. 
    
    Fig.~\ref{fig_vary_inout} shows the optimal meshes and the values of the quality metrics as a function of $x_{in}$ and $x_{out}$ where $\bf{c}$ is defined as follows:
    \begin{equation}
    \begin{gathered}
    \bf{c} = (\bf{BSP}_{\text{LS89}},~ pitch_{\text{LS89}},~ x_{in},~ x_{out},\\ 
    N_{o} = 30000,~ \Delta n_{1} = 1.1\times10^{-4}C),
    \label{eq_vary_inout}
    \end{gathered}
    \end{equation}
    where $x_{in}$ and $x_{out}$ vary from $0.2C$ to $0.8C$. For different combinations of $x_{in}$ and $x_{out}$, $(\mathcal{Q}_\mathcal{J})\vert_{min}$, $(\mathcal{Q}_\mathcal{J})\vert_{avg}$, and $(\mathcal{Q}_\mathcal{S})\vert_{avg}$ vary by $3.23\%$, $1.09\%$, and $1.20\%$, respectively, relative to their maximum values. $(\mathcal{Q}_\mathcal{S})\vert_{min}$ exhibits the largest variation of $10.00\%$ with a value of $0.40$ for the case of $0.8C$ and $0.36$ for the case of $0.2C$. This can be attributed to the fixed $pitch$ of the mesh, whereas $x_{in}$ and $x_{out}$ are shortened, leading to distortion of the cells owing to the insufficient horizontal length of the domain. The lack of horizontal length is also observed in the actions of the network. As $x_{in}$ and $x_{out}$ increase from $0.2C$ to $0.6C$, the network tries to expand the O-type mesh in the horizontal direction by moving the interfaces between the O-type mesh and the H-type meshes away from the blade. However, when $x_{in}$ and $x_{out}$ are increased from $0.6C$ to $0.8C$, the locations of the interfaces show minor changes. This suggests that the trained network attempts to locate the interfaces such that a consistent distance between each outer boundary of the O-type mesh and the blade surface is maintained, as the O-type mesh is generated as a single-block structured mesh. Moreover, it can be inferred that generating meshes with $x_{in}$ and $x_{out}$ greater than $0.8C$ can be achieved by adding extra H-type meshes at the front and the back sections.

    In addition to the blade profile, the number of blades should also be taken into account during the design process, as it considerably impacts performance, especially in applications like turbomachinery. To evaluate the effectiveness of the developed mesh generator, meshes are generated as a function of $pitch$, with $\bf{c}$ defined as follows:
    \begin{equation}
    \begin{gathered}
    \bf{c} = (\bf{BSP}_{\text{T106A}},~ pitch,~ x_{in} = 0.5C,~ x_{out} = 0.5C,\\ 
    N_{o} = 30000,~ \Delta n_{1} = 1.1\times10^{-4}C),
    \label{eq_vary_pitch}
    \end{gathered}
    \end{equation}
    where $pitch$ varies from $0.3C$ to $0.9C$. As illustrated in Fig.~\ref{fig_vary_pitch}, for different values of $pitch$, $(\mathcal{Q}_\mathcal{J})\vert_{min}$, $(\mathcal{Q}_\mathcal{J})\vert_{avg}$, $(\mathcal{Q}_\mathcal{S})\vert_{min}$, and $(\mathcal{Q}_\mathcal{S})\vert_{avg}$ vary by $3.23\%$, $2.17\%$, $6.38\%$, and $3.49\%$, respectively, relative to their maximum values. Note that the minimum values of $(\mathcal{Q}_\mathcal{S})\vert_{min}$ and $(\mathcal{Q}_\mathcal{S})\vert_{avg}$ are both observed when $pitch = 0.3C$, with values of $0.44$ and $0.83$, respectively. This is because, as $pitch$ decreases, the network increases the curvatures of the periodic boundaries to fit the mesh inside the reduced domain. Furthermore, as $pitch$ decreases, the network positions the interfaces between the O-type mesh and the H-type meshes closer to the blade surface to attain a consistent distance between each outer boundary of the O-type mesh and the blade surface.

    When simulations are required under varying flow conditions for a fixed blade, as in performance evaluations, generating meshes with different resolutions is necessary. To identify this capability, meshes are generated as a function of $N_{o}$, with $\bf{c}$ defined as follows:
    \begin{equation}
    \begin{gathered}
    \bf{c} = (\bf{BSP}_{\text{STD10}},~ pitch_{\text{STD10}},~ x_{in} = 0.5C,~ x_{out} = 0.5C,\\
    N_{o},~ \Delta n_{1} = 1.1\times10^{-4}C),
    \label{eq_vary_No}
    \end{gathered}
    \end{equation}
    where $N_{o}$ varies from $10000$ to $50000$. In Fig.~\ref{fig_vary_No}, a gradual increase in overall mesh resolution is observed as $N_{o}$ increases. For different values of $N_{o}$, $(\mathcal{Q}_\mathcal{S})\vert_{avg}$ varies by $2.35\%$, while $(\mathcal{Q}_\mathcal{J})\vert_{min}$, $(\mathcal{Q}_\mathcal{J})\vert_{avg}$, and $(\mathcal{Q}_\mathcal{S})\vert_{min}$ exhibit larger variations of $13.85\%$, $8.60\%$, and $11.11\%$, respectively, to their respective maximum values. $(\mathcal{Q}_\mathcal{J})\vert_{avg}$ gradually decreases from $0.93$ to $0.85$ as $N_{o}$ decreases. This can be attributed to that the first cell height at the blade surface is fixed at $\Delta n_{1} = 1.1\times10^{-4}C$, whereas $N_{o}$ decreases. The expansion ratio of the distribution of the nodes to the normal direction near the blade surface increases to satisfy the specified first cell height with a limited number of nodes, resulting in increased resolution changes across cells. On the other hand, $(\mathcal{Q}_\mathcal{J})\vert_{min}$ gradually decreases from $0.65$ to $0.56$ as $N_{o}$ increases. This is due to the presence of the sharp trailing edge with a high curvature. As $N_{o}$ increases, the mesh generator captures more detail in the geometry, introducing a higher degree of curvature and thus becoming more likely to produce cells with rapid resolution change or distortion. For all cases, the cells with $(\mathcal{Q}_\mathcal{S})\vert_{min}$ are located in the vicinity of the trailing edges, with minimum values of $0.40$ when $N_{o} = 10000$ and $N_{o} = 50000$. Hence, increasing the number of cells does not always guarantee higher mesh quality, and considering the minimum value of the quality metrics is essential, especially when sharp edges exist.
   
    Fig.~\ref{fig_arb} shows optimal meshes and values of the quality metrics from arbitrary sampled conditions. As shown in Fig.~\ref{fig_arb}, the developed mesh generator is capable of handling blade shapes with different thickness distributions, curvatures of the camber lines, stagger angles, and radii of the curvatures at the leading and trailing edges, assuring its robustness. This feature is beneficial in the shape optimization process where automation of simulations for various shapes is required in the optimization loop, preventing the loop from terminating owing to the introduction of unexpected shapes. Moreover, in practical usage, the blade shape is often represented by a set of scattered point coordinates rather than shape parameters such as $\bf{BSP}$. The parametrization method proposed by Agromayor \textit{et al.}~\cite{AGROMAYOR2021} can extract shape parameters $\bf{BSP}$ from the point coordinates of the blade. As the present approach employed the corresponding parametrization method, the present mesh generator can handle blade shapes represented by point coordinates, which significantly enhances the practicability of the method.

	\section{Concluding remarks} \label{sec5}
	A DRL-based mesh-generation method has been developed to non-iteratively generate an optimal mesh for various blade passages. The developed method utilizes a DRL-based MC optimization technique to determine the optimal meshing parameters as a function of the blade geometry, mitigating the human intervention and inefficiency of the conventional approaches in the mesh generation. In detail, the elliptic mesh generator to generate a structured mesh for a blade passage has been developed. Then, the MC optimization problem has been formulated to optimize the meshing parameters in the elliptic mesh generator for various blade geometries. Finally, the DRL-based mesh generator has been developed by solving the MC optimization problem using DRL. The capability of the trained mesh generator for optimal mesh generation has been identified by comparing the quality of meshes acquired in a single try with that by iterative optimization from the scratch. The practical applicability of the developed method has been confirmed by generating meshes in a single trial for a wide range of blade passages.
    
	The present method shows an outstanding performance in the mesh generation for a blade passage in the perspective of \textit{a priori} mesh quality. As one of the future directions of the present research, integration of the present method and a computational fluid dynamics simulation technique is under development to reflect \textit{a posteriori} mesh quality into optimization. Although the present study targets mesh generation for a blade passage, the proposed approach is also expected to be applicable for generation of optimal meshes for other geometric configurations~\cite{costenoble2022automated, lu2020nnw, gargallo2018mesh, zhang20182d, marchandise2013cardiovascular, zheleznyakova2013molecular}. 

	\section*{CRediT authorship contribution statement}
	\textbf{Innyoung Kim:} Conceptualization, Investigation, Methodology, Software, Validation, Visualization, Writing - Original Draft. \textbf{Sejin Kim:} Conceptualization, Investigation, Methodology, Software, Validation. \textbf{Donghyun You:} Conceptualization, Funding Acquisition, Supervision, Writing - Original Draft.
	
    \section*{Declaration of competing interest}
	The authors declare that they have no known competing financial interests or personal relationships that could have appeared to influence the work reported in this paper.
    
	\section*{Acknowledgements}
	The work was supported by the National Research Foundation of Korea (NRF) under the Grant Number NRF-2021R1A2C2092146 and the Samsung Research Funding Center of Samsung Electronics under Project Number SRFC-TB1703-51.
	
	\newpage
	%\section*{References}
    \bibliographystyle{elsarticle-num}
    \biboptions{sort&compress}
    \bibliography{DRL_Mesh}
	
	%% Algorithms %%
	\newpage
	\listofalgorithms
	
	\pagebreak
	\clearpage
	
    \begin{algorithm}[H]
    {\footnotesize
        \label{alg_drl}
	    \caption{DRL-based mesh-generation method}
		initialize actor network $\bf{\pi}_{\bf{\phi}}$ and critic network $Q_{\bf{\zeta}}$ with random parameters $\bf{\phi}$, $\bf{\zeta}$;\\
		initialize episode and replay buffer $\mathcal{B}$;\\
		\Repeat{convergence}{
		    episode $\gets$ episode+1;\\
    		randomly sample $\bf{c} = (\bf{BSP}, pitch, x_{in}, x_{out}, N_{o}, \Delta n_{1})$ from $\bf{\Phi}$;\\
    		receive state $\bf{s}$ by scaling $\bf{c}$;\\
    		select action with exploration noise: $\bf{a}\gets$clip$(\bf{\pi}_{\bf{\phi}}(\bf{s})+\bf{\epsilon},-1,1)$, $\bf{\epsilon}\sim\mathcal{N}(0,\sigma^{2})$;\\
    		determine $\bf{x} = (y_{in},y_{out},\alpha_{camber},\beta^{o}_{in},\beta^{o}_{out},N_{t},\gamma_{le},\gamma_{te})$ by rescaling $\bf{a}$;\\ 
    		generate mesh and evaluate $\mathcal{Q}$;\\
    		receive reward: $r \gets \mathcal{Q}$;\\
            store data of $(\bf{s},\bf{a},r)$ in $\mathcal{B}$;\\

        	sample mini-batch of $N_b$ data from $\mathcal{B}$;\\
        	update $\bf{\zeta}$ with the loss $N_b^{-1}\sum(r-Q_{\bf{\zeta}}(\bf{s},\bf{a}))^2$;\\			
        	\If{\textnormal{episode} mod $2$}{
            	update $\bf{\phi}$ by the deterministic policy gradient $N_b^{-1}\sum\nabla_{\bf{a}} Q_{\bf{\zeta}}(\bf{s},\bf{a})|_{\bf{a}=\bf{\pi}_{\bf{\phi}}(\bf{s})}\nabla_{\bf{\phi}}\bf{\pi}_{\bf{\phi}}(\bf{s})$;\\
        		}
        	
		}
	}
    \end{algorithm}
	
	%% Tables %%
	\newpage
	\listoftables
	
	\pagebreak
	\clearpage
	
	\begin{table}
    \caption{Parameters required for the elliptic mesh generator and their descriptions.}
    \begin{center}
    \begin{footnotesize}
    \begin{tabularx}{\textwidth}{ l l } 
        \hline
        Geometric parameter & Description \\
        \hline
        Blade shape & Set of scattered point coordinates of the blade  \\[-0.5pt]
        $pitch$ & Blade spacing  \\[-0.5pt]
        $x_{in}$ & Inlet position in the horizontal direction  \\[-0.5pt]
        $x_{out}$ & Outlet position in the horizontal direction  \\ \hline

        Meshing parameter & Description \\
        \hline
        $y_{in}$ & Inlet position in the vertical direction  \\[-0.5pt]
        $y_{out}$ & Outlet position in the vertical direction  \\[-0.5pt]
        $\alpha_{camber}$ & Degree of the curvature of the lower boundary\\[-8pt] 
        & following the camber line  \\[-0.5pt]
        $\beta^{o}_{in}$ & HO-type interface position at the inlet normalized by $x_{in}$ \\[-0.5pt]
        $\beta^{o}_{out}$ & OH-type interface position at the outlet normalized by $x_{out}$ \\[-0.5pt]
        $N_{t}$ & Number of nodes of the O-type mesh\\[-8pt] 
        & in the tangential direction to the blade surface  \\[-0.5pt]
        $N_{n}$ & Number of nodes of the O-type mesh\\[-8pt] 
        & in the normal direction to the blade surface  \\[-0.5pt]
        $\gamma_{le}$ & Degree of clustering of nodes at the leading edge  \\[-0.5pt]
        $\gamma_{te}$ & Degree of clustering of nodes at the trailing edge  \\[-0.5pt]
        $\Delta n_{1}$ & First cell height in the direction normal to the blade surface \\
        \hline
    \end{tabularx}
    \end{footnotesize}
    \end{center}
    \label{table_params}
    \end{table}
    
    \pagebreak
    \clearpage
    
    \begin{table}
    \caption{Mesh quality $\mathcal{Q}$ non-iteratively obtained by the trained mesh generator using deep reinforcement learning for the test conditions ($\bf{c}_{\text{LS89}}$, $\bf{c}_{\text{STD10}}$, $\bf{c}_{\text{T106A}}$, and $\bf{c}_{\text{SIRT}}$). $\mathcal{Q}$ consists of the quality metrics $\mathcal{Q}_{\mathcal{J}}$ and $\mathcal{Q}_{\mathcal{S}}$, which denote the determinant ratio of the Jacobian matrix and the skewness, respectively. $(~)\vert_{min}$ and $(~)\vert_{avg}$ denote the minimum and average values, respectively.}
    \begin{center}
    \begin{footnotesize}
    \begin{tabular}{ c c c c c c} 
        \hline
        Condition & $(\mathcal{Q}_\mathcal{J})\vert_{min}$ & $(\mathcal{Q}_\mathcal{J})\vert_{avg}$ & $(\mathcal{Q}_\mathcal{S})\vert_{min}$ & $(\mathcal{Q}_\mathcal{S})\vert_{avg}$ & $\mathcal{Q}$\\
        \hline
        $\bf{c}_{\text{LS89}}$ & 0.62 & 0.92 & 0.37 & 0.83 & 0.69\\
        $\bf{c}_{\text{STD10}}$ & 0.58 & 0.91 & 0.47 & 0.83 & 0.70\\
        $\bf{c}_{\text{T106A}}$ & 0.62 & 0.90 & 0.47 & 0.86 & 0.71\\
        $\bf{c}_{\text{SIRT}}$ & 0.62 & 0.91 & 0.49 & 0.80 & 0.71\\
        \hline
    \end{tabular}
    \end{footnotesize}
    \end{center}
    \label{table_testset_quality}
    \end{table}
    
    \pagebreak
    \clearpage
	
	%% Figures %%
	\newpage
	\listoffigures
	
	\pagebreak
	\clearpage
	
	\pagebreak
	\clearpage
	
	\pagebreak
	\clearpage

    \pagebreak
	\clearpage

	\pagebreak
	\clearpage

    \pagebreak
	\clearpage
	
	\pagebreak
	\clearpage

    \pagebreak
    \clearpage

    \pagebreak
    \clearpage
	
	\pagebreak
    \clearpage
	
    \pagebreak
    \clearpage
	
\pagebreak
\clearpage

\end{document}